%% file: PaperForReview.tex
\pdfoutput=1
\documentclass[10pt,twocolumn,letterpaper]{article}

\usepackage[pagenumbers]{cvpr} 

\usepackage{graphicx}
\usepackage{amsmath}
\usepackage{amssymb}
\usepackage{booktabs}
\usepackage{framed}
\usepackage{enumitem}

 \usepackage{array, makecell}
 \usepackage{boldline}
\usepackage{multirow}
 \usepackage{amsmath}

\usepackage[pagebackref,breaklinks,colorlinks]{hyperref}
\usepackage[dvipsnames]{xcolor}

\usepackage[capitalize]{cleveref}
\crefname{section}{Sec.}{Secs.}
\Crefname{section}{Section}{Sections}
\Crefname{table}{Table}{Tables}
\crefname{table}{Tab.}{Tabs.}
\newcommand{\inctabcolsep}[2]{\addtolength{\tabcolsep}{#1} #2 \addtolength{\tabcolsep}{-#1}}

\def\cvprPaperID{4346} 
\def\confName{CVPR}
\def\confYear{2022}

\begin{document}

\title{ZeroWaste Dataset: \\ Towards Deformable Object Segmentation in Cluttered Scenes}
  
  
  
  
  
  


\author{Dina Bashkirova$^{1}$\thanks{dbash@bu.edu}\\
\and
  Mohamed Abdelfattah$^{2}$
  
  \and
  Ziliang Zhu $^{1}$\\

  \and
  James Akl$^{3}$ \\

  \and
  Fadi Alladkani  $^{3}$
  \and
  Ping Hu $^{1}$\\
  
  \and
  Vitaly Ablavsky $^{4}$
  
  \and
  Berk Calli $^{3}$\\
  
  \and
  Sarah Adel Bargal$^{1}$\\

\and
Kate Saenko$^{1,5}$\\

\and
$^{1}$ Boston University \ \ \ $^{2}$American University in Cairo \ \ \ $^{3}$ Worcester Polytechnic Institute \and \ \ \ $^{4}$ University of Washington \ \ \ $^{5}$ MIT-IBM Watson AI Lab
}

\def\eg{\emph{e.g. }}
\def\ie{\emph{i.e.} }
\def\etal{\emph{et al.} }
\def\etc{\emph{etc. }}
\def\zerowaste{{\fontfamily{qcr}\selectfont ZeroWaste}}
\def\zerowastef{{\fontfamily{qcr}\selectfont ZeroWaste-\textit{f}}}
\def\zerowastew{{\fontfamily{qcr}\selectfont ZeroWaste-\textit{w}}}
\def\zerowastes{{\fontfamily{qcr}\selectfont ZeroWaste-\textit{s}}}
\def\zerowasteaug{{\fontfamily{qcr}\selectfont ZeroWaste\textit{Aug}}}

\maketitle

\begin{abstract}
  Less than 35\% of recyclable waste is being actually recycled in the US~\cite{waste_stats}, which leads to increased soil and sea pollution and is one of the major concerns of environmental researchers as well as the common public. At the heart of the problem are the inefficiencies of the waste sorting process (separating paper, plastic, metal, glass, etc.) due to the extremely complex and cluttered nature of the waste stream. 
{\color{black} Recyclable waste detection poses a unique computer vision challenge as it requires detection of highly deformable and often translucent objects in cluttered scenes without the kind of context information usually present in human-centric datasets}.
  This challenging computer vision task currently lacks suitable datasets or methods in the available literature. In this paper, we take a step towards computer-aided waste detection and present the first in-the-wild industrial-grade waste detection and segmentation dataset, \zerowaste. 
 We believe that \zerowaste~will
 catalyze research in object detection and semantic segmentation in extreme clutter as well as applications in the recycling domain.
Our project page can be found at \url{http://ai.bu.edu/zerowaste/}
\end{abstract}

\input{1_intro}

\input{2_related_work}

\input{3_dataset}

\input{4_experiments}

\input{5_discussion}

{\small
\bibliographystyle{ieee_fullname}
\bibliography{egbib}
}

\input{ 6_supplementary}

\end{document}

%% file: 1_intro.tex
\section{Introduction}
\label{sec:intro}

As the world population grows and becomes increasingly urbanized, waste production is estimated to reach 2.6 billion tonnes a year in 2030, an increase from its current level of around 2.1 billion tonnes \cite{World_Bank_2018}. Efficient recycling strategies are critical to reduce the devastating environmental effects of rising waste production. Materials Recovery Facilities (MRFs) are at the center of the recycling process. These facilities are where the collected recyclable waste is sorted into separate bales of plastic, paper, metal and glass (and other sub-categories). 
Even though the MRFs utilize a large number of machines alongside manual labor \cite{gundupalli2017review}, the recycling rates as well as the profit margins stay at undesirably low levels (e.g. less than 35\% of the recyclable waste actually got recycled in the United States in 2018~\cite{waste_stats}). Another crucial aspect of manual waste sorting is the safety of the workers, who risk their health daily getting exposed to dangerous and unsanitary objects (\eg knifes, medical needles). At the same time, the extremely cluttered nature of the waste stream makes automated waste detection (\ie detection of waste objects that should be removed from the conveyor belt) very challenging to achieve.


\begin{figure*}[htp]
    \centering
    \includegraphics[width=\linewidth]{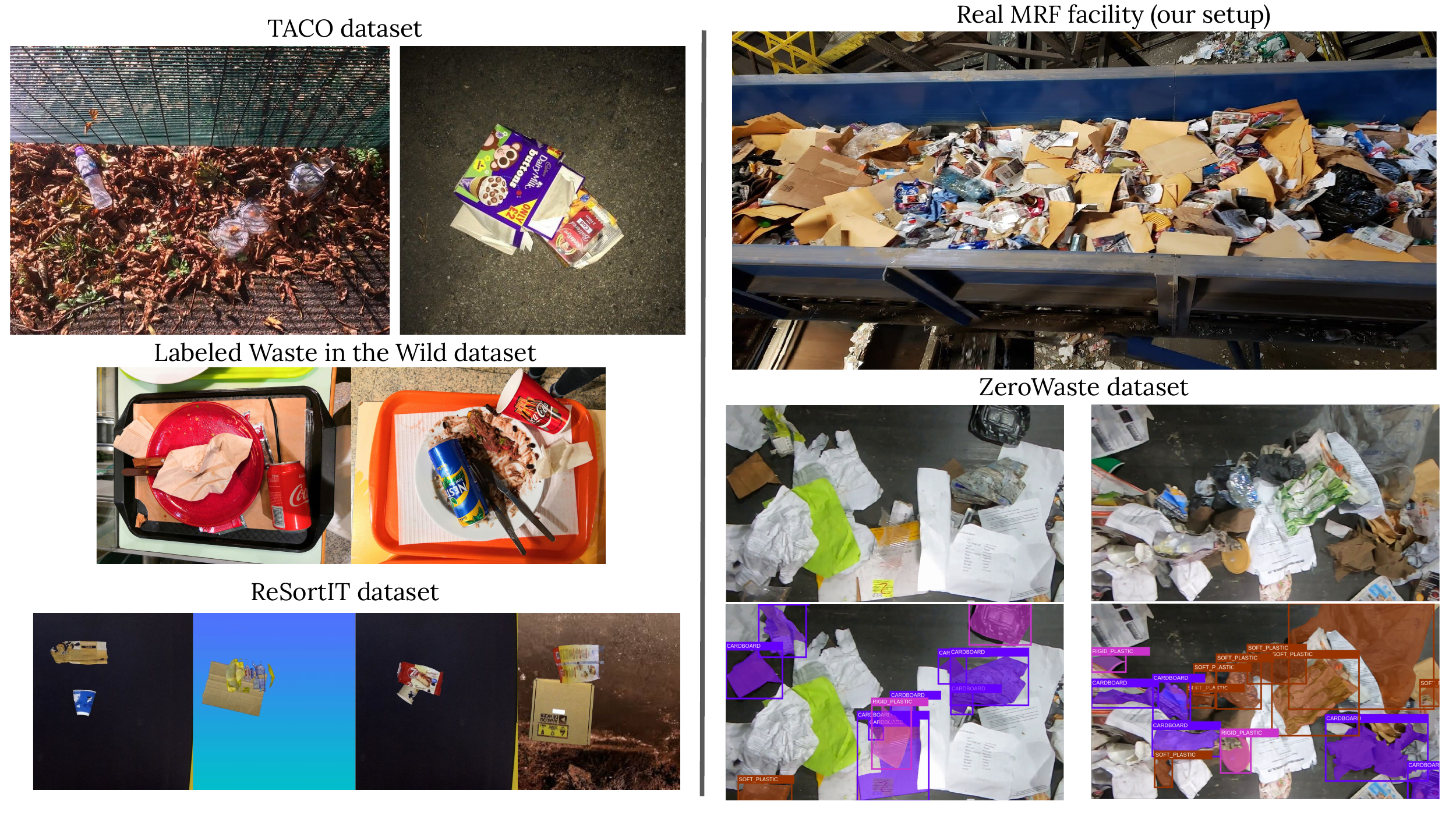}
    \caption{{\textbf{Left:} examples of the existing waste detection and classification datasets (top to bottom): Trash Annotation in Context (TACO)~\cite{proencca2020taco}, Labeled Waste in the Wild~\cite{sousa2019automation}, ReSortIT~\cite{koskinopoulou2021robotic} datasets. \textbf{Right:} footage of the waste sorting process at a real Materials Recovery Facilities (MRF). The domain shift between the simplified datasets with solid background and little to no clutter
     and the real images of the conveyor belt from the MRF, as well as the object-centric nature of the existing datasets,
     makes it impossible to use models trained on these datasets for automated detection on real waste processing plants. In this paper, we propose a new \zerowaste~dataset collected from a real waste sorting plant. Our dataset includes a set of densely annotated frames for training and evaluation of the detection and segmentation models, as well as a large number of unlabeled frames for semi- and self-supervised learning methods. We also include frames of the conveyor belt before and after manual collection of foreground objects to facilitate research on weakly supervised detection and segmentation. Please see Figure~\ref{fig:fully_annotated_example} for the illustration of our \zerowaste~dataset.}}

    \label{fig:mrf}
\end{figure*}

Recent advances in object classification and segmentation provide a great opportunity to make the recycling process more efficient, more profitable and safer for the workers. %
Unfortunately, the research community is lacking the high-quality in-the-wild datasets to train and evaluate the classification and segmentation algorithms for industrial waste sorting. While several companies do development in the automated waste sorting space (e.g.~\cite{AMP-Robotics,Waste-Robotics,Zen-Robotics}), they keep their data private, and the few existing open-source datasets~\cite{yang2016classification, PSUWaste, sousa2019automation, proencca2020taco} are very limited in the amount of data and/or are generated in uncluttered environments, not representing the complexity of the domain (see Figure~\ref{fig:mrf}).

In this paper, we propose the largest openly available in-the-wild waste detection dataset \zerowaste~that is specifically designed for evaluating label-efficient industrial waste detection. \zerowaste~is a dataset that is fundamentally different from the popular detection and segmentation benchmarks: high level of clutter, presence of highly deformable and translucent objects, as well as a fine-grained difference between the object classes -- all these aspects pose a unique challenge for the automated vision. In addition to that, due to the ever-changing nature of the stream, content and visual qualities of the stream are often MRF-specific and highly depend on the season, therefore the detection algorithm must be label-efficient and able to learn and adapt to the changes in the stream with only a few labeled examples. 
We envision that our open-access dataset will allow researchers to develop more robust and data-efficient algorithms for object detection and other related problems beyond human-centric domains. 
\begin{figure*}[ht]
    \centering
    \includegraphics[width=1.\linewidth]{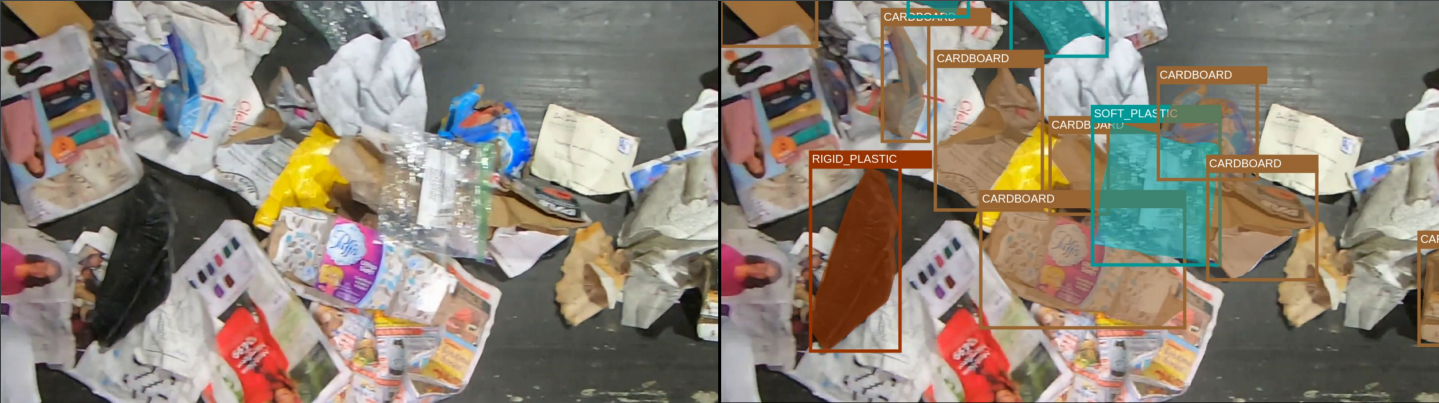}
    \caption{{\textbf{Left:} example of an image from \zerowastef~dataset. \textbf{Right:} the corresponding ground truth instance segmentation. At the end of this conveyor belt, only paper objects must remain. Therefore, we annotated the removable objects of four material types as foreground: soft plastic, rigid plastic, cardboard and metal. The background includes the conveyor belt and paper objects. Severe clutter and occlusions, high variability of the foreground object shapes and textures, as well as severe deformations of objects usually not present in other segmentation datasets, make this domain very challenging for object detection. More examples of our annotated data can be found in Section~\ref{sec:more_examples} of the Appendix
    (\textit{best viewed in color}). }}
    \label{fig:fully_annotated_example}
\end{figure*}
We summarize our contributions as follows: 
\begin{enumerate}[leftmargin=*]
    \item We propose the first fully-annotated \zerowastef~dataset industrial waste object detection. The \zerowastef~dataset presents a challenging real-life computer vision problem of detecting highly deformable objects in severely cluttered scenes. In addition to the fully annotated frames from \zerowastef~set, we include the unlabeled \zerowastes~set for semi-supervised learning. We also propose a version of our \zerowaste~data augmented with objects from the TACO~\cite{proencca2020taco} dataset, \zerowasteaug, to combat class imbalance. We show that introduction of object augmentation improves the overall segmentation quality.
    
    \item We introduce a novel before-after data collection setup and propose the \zerowastew~dataset for binary classification of frames before and after the collection of target objects. This binary classification setup allows much cheaper data annotation and catalyzes further development of weakly supervised segmentation and detection methods. {\color{black} Our experimental results show that meaningful foreground segmentation can be achieved using \zerowastew, however, more efficient weakly-supervised methods are needed to reach the segmentation quality achieved by fully-supervised methods.}
    
    \item We implement the fully-supervised detection and segmentation baselines for the \zerowastef~dataset and semi- and weakly-supervised baselines for \zerowastes~and \zerowastew~datasets. Our results show that popular detection and segmentation methods, such as Mask-RCNN, TridentNet and DeepLabV3+, struggle to generalize to our data, which indicates a challenging nature of our in-the-wild dataset and suggests that new and more robust methods must be developed to solve the problem efficiently and be applied in the real waste sorting plants.
\end{enumerate}

%% file: 2_related_work.tex
\section{Related work}
\label{sec:related_work}
\paragraph{Detection and Segmentation Datasets}
Many datasets for image segmentation have been proposed with the goal of densely recognizing general objects and ``stuff'' in  image scenes like street view~\cite{cordts2016cityscapes,yu2020bdd100k,brostow2008segmentation}, natural scenes~\cite{Everingham10,caesar2018coco,mottaghi2014role,zhou2017scene,lin2014microsoft}, and indoor spaces~\cite{silberman2012indoor,dai2017scannet,chen2018encoder}. 
Yet, few of them have been designed for the more challenging vision task required in automated waste recycling.
 Several related datasets have been proposed that contain only image-level labels, like \textit{Portland State University Recycling}~\cite{PSUWaste} consists of labeled images of box-board, glass bottles, soda cans, crushed soda cans and plastic bottles, and a small-scale \textit{Stanford TrashNet}~\cite{yang2016classification} dataset containing images of single waste objects from six predefined classes. 
Though beneficial for image-level classification in well-defined conditions, images of in these two datasets have very simple background and do not apply to waste object localization. 
To enable localization tasks, \textit{Labeled Waste in the Wild}~\cite{sousa2019automation} annotated bounding boxes for objects of 20 waste classes.  Perhaps the most similar dataset to ours is the
\textit{Trash Annotation in Context (TACO)}~\cite{proencca2020taco} of 1500 densely annotated images containing objects of 60 litter classes. 
Yet TACO contains deliberately collected outdoor scenes with one or a few foreground objects that are rarely occluded, which makes it less practical for materials recovery scenarios (see Table~\ref{tab:detection_stats} and Figure~\ref{fig:mrf} for comparison). {\color{black} Another great effort towards automated waste sorting is the synthetic ReSort-IT~\cite{koskinopoulou2021robotic} dataset for waste object detection on the conveyor belt, as well as the FloW~\cite{cheng2021flow} dataset of the floating waste detection in the inland waters.}
In contrast, our \zerowaste~ has almost $3$ times more annotated images and more than $10$ times more annotated instances than TACO and is collected from the front lines of a waste sorting plant
where the objects are frequently severely deformed and occluded, which makes our dataset the closest one to the real applications, \eg, robotic grasping or waste stream analysis. 

\paragraph{Detection and Segmentation Methods} 
Image segmentation can be formulated as a task of classifying each pixel into a set of labels~\cite{minaee2021image}. 
Recent semantic segmentation models~\cite{zhao2017pyramid,chen2017rethinking,yuan2019object,liu2021Swin} have achieved state-of-the-art performance for recognizing general object/stuff classes from natural scene images.
Representative frameworks like MaskRCNN~\cite{massa2018mrcnn} effectively detect objects in images and simultaneously generate high-quality masks, which enables efficient interaction between robots and target objects. 
Yet due to their data-hungry nature, these methods rely on large volumes of annotated data for training, which can be challenging and expensive, especially in specialized application scenarios~\cite{barriuso2012notes}. Recycling annotation in particular requires expert labelers and is thus even more costly. 
Semi-supervised segmentation methods have been proposed to address such limitations by jointly learning from both annotated and unannotated images~\cite{Ouali_2020_CVPR,mittal2019semi,mendel2020semi, kim2020structured,french2019semi,chen2020semiVS}. 
For example, ReCo~\cite{reco_ref} introduces an unsupervised loss to optimize both intra- and inter-class variance of pixels so as to further boost the segmentation performance.
In a more data-efficient setting, weakly-supervised segmentation methods exploit annotations that are even easier to obtain, e.g. image-level tags~\cite{ahn2018learning,kolesnikov2016seed,pathak2015constrained}. 
These methods typically utilize Class Activation Maps (CAM)~\cite{zhou2016learning} to select the most discriminative regions, which are later used as pixel-level supervision for segmentation networks~\cite{Chang_2020_CVPR,Fan_2020_CVPR,Wang_2020_CVPR}. 
By constraining consistency between partial and full features,  PuzzleCAM~\cite{puzzle_ref} effectively enhances the quality of CAMs without adding layers.
All these advanced segmentation models are trained on large-scale general-purpose data, 
such as MS-COCO~\cite{lin2014microsoft} or PASCAL VOC~\cite{Everingham10}, and applying them to the cluttered real-world scenarios presents challenges like domain shift and poor generalization. We provide the experimental results on the popular fully-, semi- and weakly-supervised semantic segmentation and object detection methods on our \zerowaste~dataset and analyze their performance on this real industrial task.


%% file: 3_dataset.tex
\begin{figure}[b]
    \centering
    \includegraphics[width=0.48\linewidth]{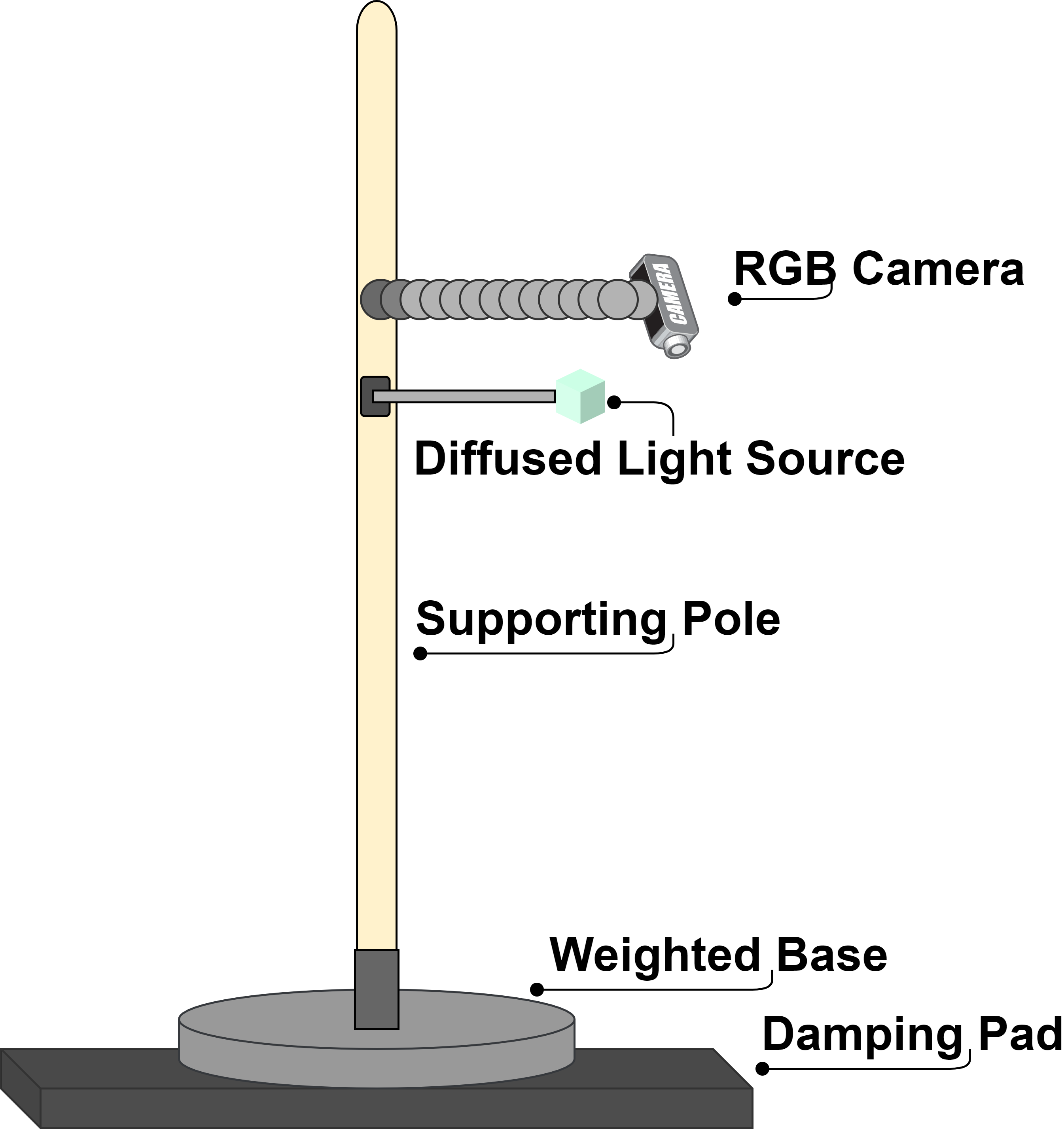}
    \includegraphics[width=0.48\linewidth]{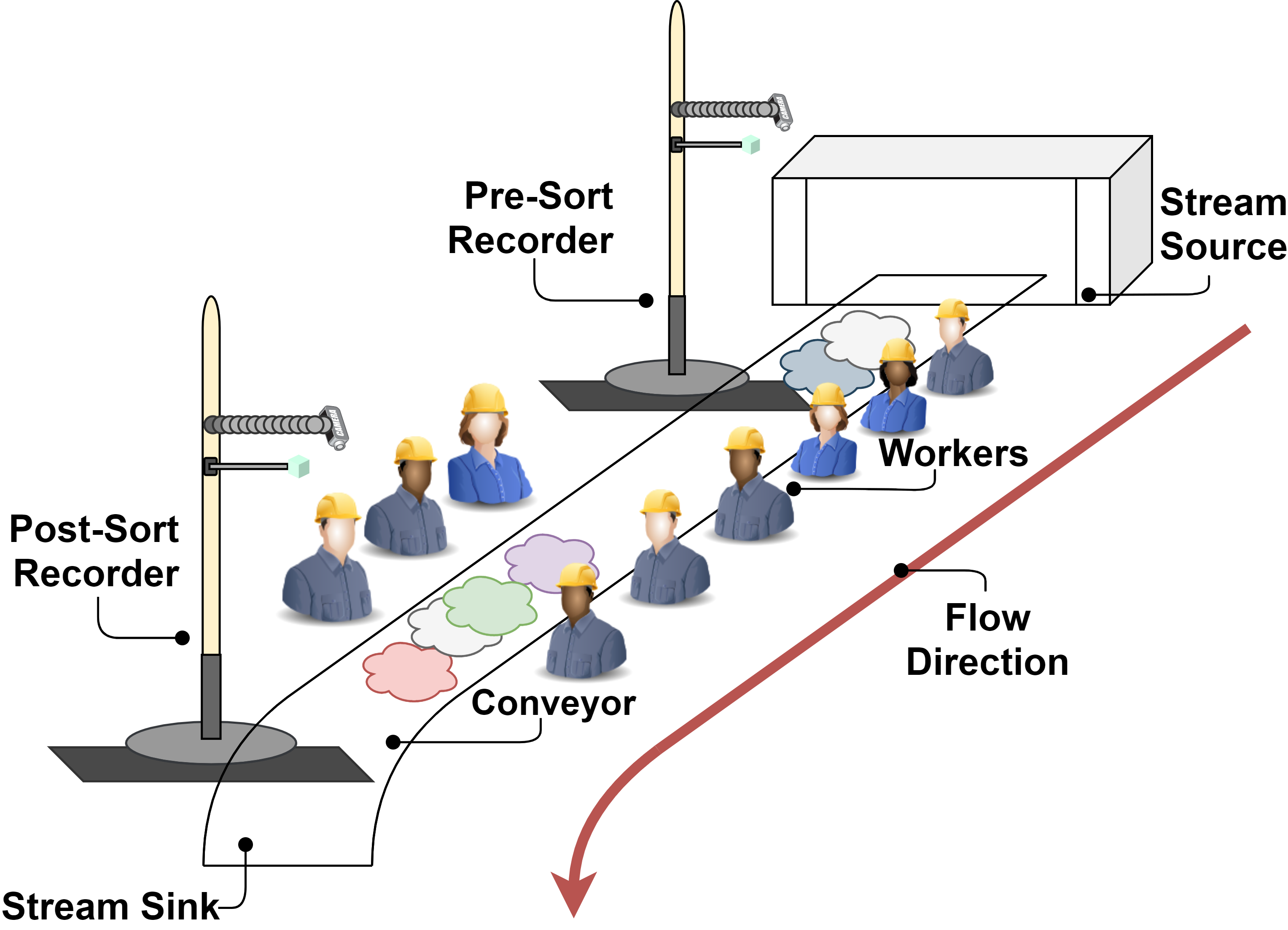}
    \caption{{The footage recording setup is designed to fit the constraints of the facility environment. \textbf{Left:} Assembly of each recording apparatus. \textbf{Right:} Layout of the recording setup in the recycling environment.}}
    \label{fig:setup}
\end{figure}
\section{\zerowaste~Dataset}
\label{sec:dataset_description}
\begin{figure}[t]
    \centering
    \includegraphics[width=0.48\textwidth]{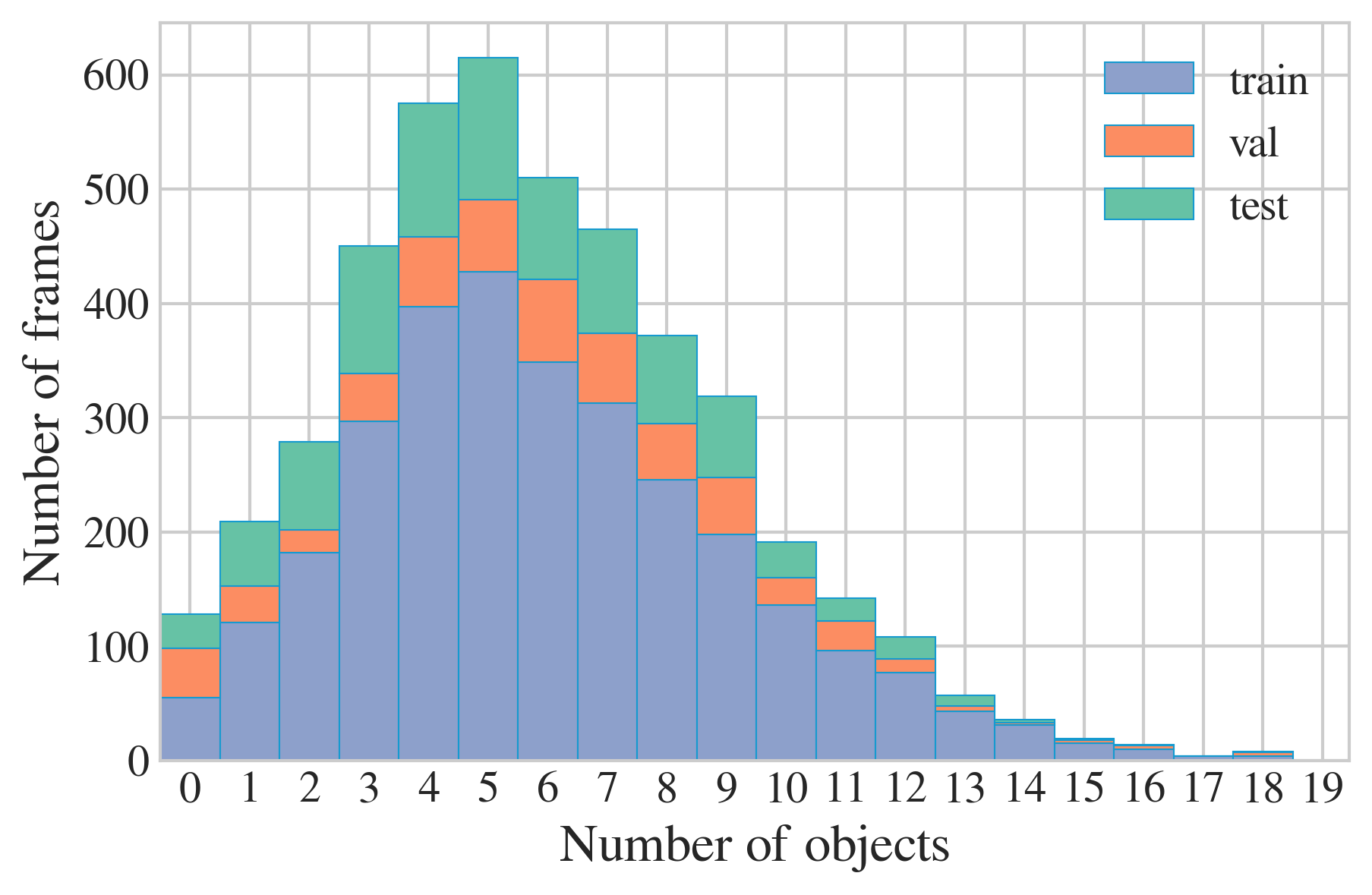}
    \caption{{Statistics of the number of \emph{foreground} objects per frame in train, val and test splits of \zerowastef~. The overall number of objects in the scene, including paper objects, is much higher (\textit{best viewed in color}).}}
    \label{fig:splits_stats}
\end{figure}
\begin{figure}[t]
    \centering
  \includegraphics[width=\linewidth, height=8cm,trim=6.5cm 3cm 6.5cm 2cm,clip]{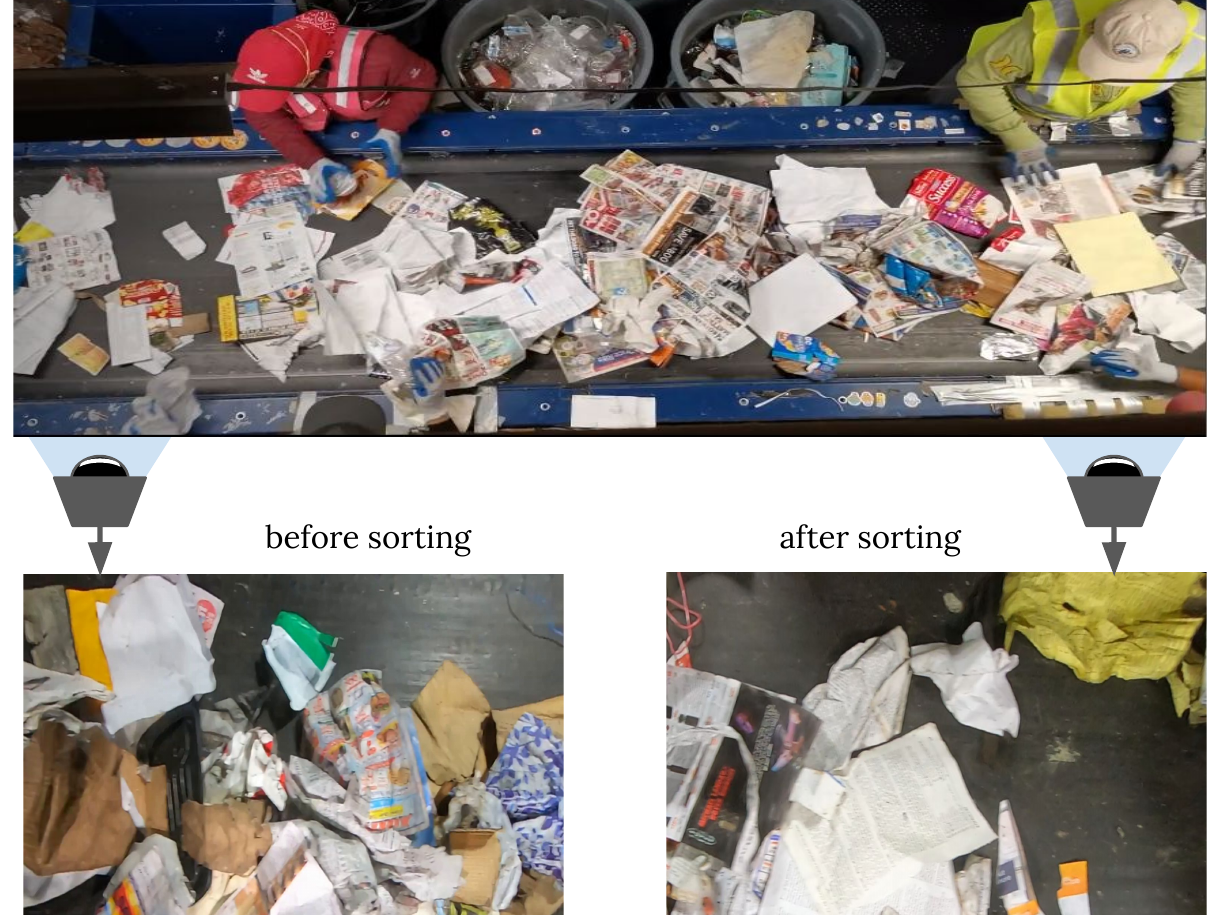}
    \caption{\footnotesize We installed two stationary cameras above the conveyor belt: one at the beginning of the line and another one at the end. At this particular conveyor belt, workers are asked to remove objects of any material other than paper, such as cardboard, plastic and metal. Therefore, the footage collected from the beginning of the line contains the ``foreground" objects that need to be removed, and the frames from the end of the conveyor belt are supposed to only contain the ``background" paper objects. We used this setup as a foundation of our \zerowastew~dataset.}
    \label{fig:data_collection_process}
\end{figure}

In this section, we describe our \zerowastef~~dataset for fully supervised detection and evaluation, unlabeled \zerowastes~for semi-supervised learning, \zerowastew~of images before and after the removal of target objects for weakly supervised detection, and \zerowasteaug~of multi-domain augmented waste objects for comparison in all fully-, semi-, and weakly-supervised settings. The datasets are licensed under the Creative Commons Attribution-NonCommercial 4.0 International License~\cite{License}. The MRF at which the data was collected agreed to release the data for any non-commercial purposes and decided to remain unacknowledged. 
\paragraph{Data Collecion and Pre-processing}
The data was collected from a high-quality paper conveyor of a single stream recycling facility in Massachusetts. The sorting operation on this conveyor aims to keep high quality paper and consider anything else as contaminants including non-paper items (\eg metal, plastic, brown paper, cardboard, boxboard). We collected data during the regular operation of the MRF using two compact recording installations at the start and end of the conveyor belt (see Fig. ~\ref{fig:data_collection_process}),  that is, footage is captured simultaneously both at the unsorted and sorted sections of the same conveyor. The recording apparatus is designed to fit the constraints of the facility: In order not to disrupt the MRF operation and be able to work in confined spaces available near the conveyor the recording platform needs to be compact, non-intrusive (to the workers), and portable (easy to move, battery-powered). Note that the cameras are not directly mounted on the conveyor but to a stand-alone platform, to reduce vibrations transmitted to the cameras. Additional considerations are made (see Figure~\ref{fig:setup}): (1) Damping pads are installed to counter the ground vibrations of the heavy machinery and reduce vibrations on the camera even further; (2) Weighted bases lower the center of mass to keep the apparatus stable.
We used the GoPro Hero 7 camera to collect the video data, and installed two LitraTorch 2.0 portable lamps with a light diffuser to maintain consistent lighting.
Both cameras were installed at around 100 cm above the conveyor, and the light sources at around 80 cm. 

Sequences of 12 videos of total length of $95$ minutes and $14$ seconds with FPS $120$ and size $1920\times1080$ were collected and processed through the following steps:

\begin{enumerate}[leftmargin=*]
    \setlength{\itemsep}{0pt}
  \setlength{\parskip}{0pt}
    \setlength{\parsep}{0pt}
    \item Rotation and cropping. The frames were rotated so that the conveyor belt is parallel to the frame borders and cropped to remove the regions outside the conveyor belt. We ensured that any personal information or identifiable footage of the workers at the conveyor belt was excluded from our data.
    \item Optical distortion. We removed the distortion~\cite{brown1966brownmodel} using the OpenCV~\cite{opencv_library} library to compensate for the fish-eye effect caused by the proximity of the cameras to the conveyor belt.
    \item Deblurring. We used the SRN-Deblur~\cite{tao2018srndeblur} method to remove motion blur resulting from the fast-moving conveyor belt. According to our visual inspection, SRN-Deblur achieves satisfactory deblurring and does not introduce the undesired artifacts that usually appear when classical deconvolution-based methods are used. 
\end{enumerate}
The illustration of the original frames shot at the beginning of the conveyor belt and the corresponding preprocessing results can be found in Figure~\ref{fig:processing} in Section~\ref{sec:more_examples} of the Appendix.
\paragraph{Densely Annotated \zerowastef~and Unlabeled \zerowastes~Datasets}

The fully annotated \zerowastef~dataset consists of $4661$ frames sampled from the processed videos and the corresponding ground truth polygon segmentation. We used the open-source CVAT~\cite{boris_sekachev_2020_4009388} annotation toolkit to manually collect the polygon annotations of objects of four material types: cardboard, soft plastic, rigid plastic and metal. We chose this set of class labels following the MRF's guidelines for the workers to collect cardboard, plastic and metal into separate bins, as well as the fact that grasping of rigid and non-rigid objects might require the use of  fundamentally different kinds of robotic systems. We annotated $1805$ frames with a subsampling rate of $10$ to enable training and evaluation of object tracking methods, as well as $2616$ frames with a subsampling rate of $100$ to increase the dataset diversity.
The polygon annotation was performed according to the following set of rules: 

\begin{figure}[b]
    \centering
    \includegraphics[width=\linewidth,trim=0cm 6cm 0cm 0cm,clip]{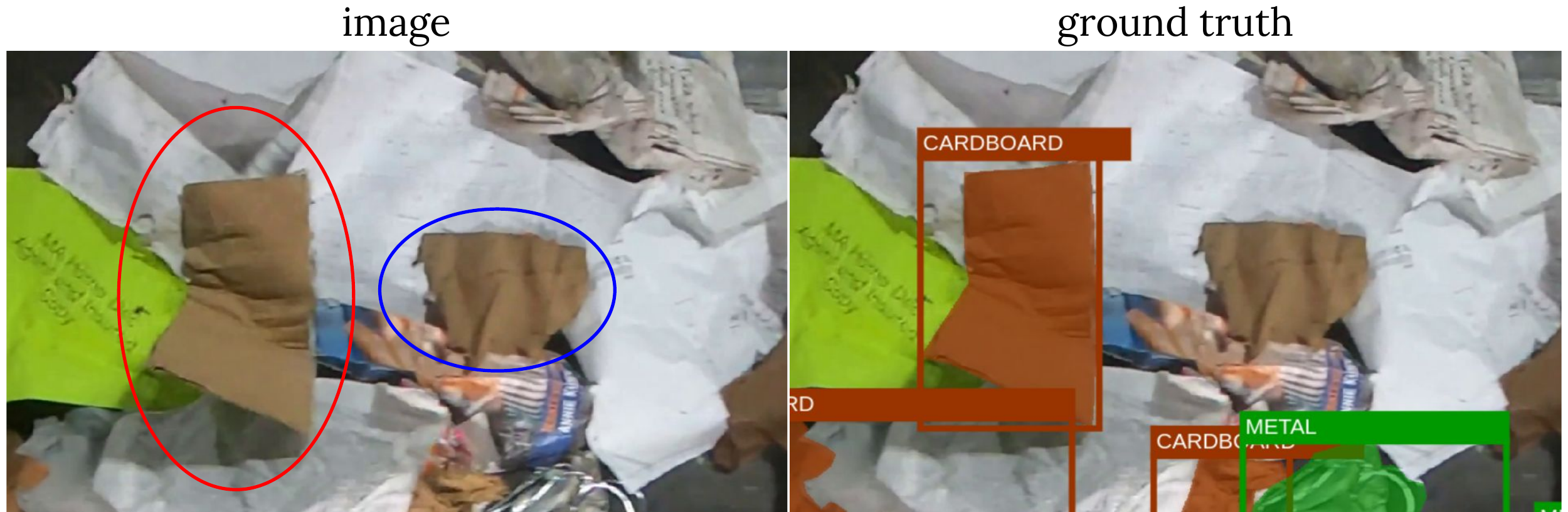}
    \caption{\footnotesize{\textbf{Left:} example of an image from \zerowastef~dataset. \textbf{Right:} the corresponding ground truth instance segmentation. Expert training and common sense knowledge are required to distinguish between the cardboard object on the left ({\color{red} red circle}) and the brown paper on the right ({\color{blue} blue circle}), as they are visually very similar but differ in thickness and rigidity. {\color{black} Trained annotators achieve average pixel-level precision of $79.57\%$ and recall of $71.96\%$ for the cardboard class.} (\textit{best viewed in color}).}}
    \label{fig:cardbaord_paper_example}
\end{figure}

\begin{enumerate}[leftmargin=*]
    \setlength{\itemsep}{0pt}
  \setlength{\parskip}{0pt}
    \setlength{\parsep}{0pt}
    \item Objects of four material types were annotated as foreground: cardboard (including parcel packages, boxboard such as cereal boxes and other carton food packaging), soft plastic (\eg plastic bags, wraps), rigid plastic (\eg food containers, plastic bottles) and metal (\eg metal cans). Paper objects were treated as background.
    \item A polygon must include all foreground object pixels, and it might include a small amount of background pixels.
    \item If an object is partially occluded and separate parts are visible, we annotated them as separate objects, since it is is impossible to certainly tell whether it is one or several visually similar objects.
\end{enumerate}
\begin{table*}[ht]\small
    \centering
    \begin{tabular}{r| c  c c c c  c  c}
        \toprule
        \textbf{Split} & \textbf{\#Images} & \textbf{\textit{Cardboard}} & \textbf{\textit{Soft Plastic}} & \textbf{\textit{Rigid Plastic}} & \textbf{\textit{Metal}} & \textbf{\#Objects} & \textbf{Domain} \\
        \midrule[0.2mm]
        \textbf{Train} &  $3002$ & $12940$ & $4862$ & $1160$ {\footnotesize\color{PineGreen} $+2778$} & $263$ {\footnotesize\color{PineGreen} $+4373$} & $19225$ {\footnotesize\color{PineGreen} $+7151$} & \multirow{6}{*}{\text{Real}} \\
        \textbf{Validation} & $572$ & $2167$ & $855$ & $305$   {\footnotesize\color{PineGreen} $+637$} & $60$  {\footnotesize\color{PineGreen} $+1010$} & $3387$ {\footnotesize\color{PineGreen} $+1647$} & \\
        \textbf{Test} & $929$ & $3428$ & $1236$ & $315$ {\footnotesize\color{PineGreen} $+886$} & $63$ {\footnotesize\color{PineGreen} $+990$}  & $5042$ {\footnotesize\color{PineGreen} $+1876$} & \\
        \textbf{Unlabeled} & $6212$ & - & - & - & - & - & \\
        \textbf{Total} & $10715$ & $18535$ & $6953$ & $1780$  {\footnotesize\color{PineGreen} $+4301$} & $386$ {\footnotesize\color{PineGreen} $+6370$} & $27744$ {\footnotesize\color{PineGreen} $+10584$} & \\
        \midrule
        \textbf{TACO} & $1499$ & $240$ & $652$ & $1183$ &  $506$ & $2581$ & Real \\
        \textbf{ReSortIT} & $16000$ & $8000$ & $8000$ & $8000$ & $8000$ & $32000$ & Synthetic\\
        \bottomrule

    \end{tabular}
    \vspace{0.5em}
    \caption{{Statistics of the total number of objects in the training, validation and test splits of \zerowasteaug~and \zerowastef~datasets, in comparison with TACO~\cite{proencca2020taco} dataset dataset with labels mapped to our set of classes and the Synthetic Complex subset of the ReSortIT~\cite{koskinopoulou2021robotic} dataset. The numbers written in {\color{PineGreen} green}, \eg {\footnotesize\color{PineGreen} $+10584$}, reflect the number of augmented objects added in \zerowasteaug. Row 4 presents the unlabeled \zerowastes~set of images for semi-supervised learning.}}
    \label{tab:detection_stats}
\end{table*}

Each annotated video frame was validated by an independent reviewer to pass the standards above (see Figure~\ref{fig:fully_annotated_example}). The review process was performed by the students and researchers with a computer science background specifically trained to perform the annotation. We did not delegate the annotation to the crowd-sourcing platforms, such as Amazon Mechanical Turk~\cite{crowston2012amazon}, due to the complexity of the domain that requires expert knowledge to be able to detect and correctly classify the foreground objects (see the illustration in Figure~\ref{fig:cardbaord_paper_example}). {\color{black} Instead, the professional annotators have been hired via Upwork\footnote{https://www.upwork.com/} to annotate the frames, with the estimated average time spent on the annotation and review about $12.5$ minutes per frame (details on that can be found in Section~\ref{sec:anno_costs} of Appendix). Each frame was assessed by an expert reviewer for consistency and quality control. The resulting average annotation agreement across $20$ frames is $84.2\%$ before expert review and above $94\%$ after the review, more details on the human annotation agreement can be found on Figure~\ref{fig:conf_mrt_main}.}
The dataset was split into training, validation and test splits and stored in the widely used MS COCO~\cite{lin2014microsoft} format for object detection and segmentation using the open-source Voxel51 toolkit~\cite{moore2020fiftyone}. Please refer to Table~\ref{tab:detection_stats} and Figure~\ref{fig:splits_stats} for more details about the class-wise statistics of all splits. 
In addition, we provide $6212$ unlabeled images that comprise \zerowastes~dataset. \zerowastes~ can be used to refine the detection using semi-supervised or self-supervised learning methods. 

\noindent \textbf{\zerowasteaug~Dataset}
As seen in Table~\ref{tab:detection_stats}, there is a significant imbalance in the class distribution of the collected ~\zerowaste~dataset that can degrade the performance of the detection methods, \eg metal objects are more than 40 times more rare than the cardboard. 
\zerowasteaug~comprises a version of \zerowastef~in which frames were augmented with the objects of the rare metal and rigid plastic classes that were cropped out of TACO~\cite{proencca2020taco} dataset.
To minimize the domain shift between the TACO objects and the \zerowaste~ frames, the contrast, brightness, and blurriness of each cropped object were changed to match the average value of the frame on which the object was augmented. Random resizing, cropping, mirroring were applied on each TACO object before augmentation (please refer to Figure~\ref{fig:more_fully_annotated_examples_1_aug} for the examples). 

\paragraph{Weakly supervised \zerowastew~Dataset}
We leverage the videos taken of the conveyor belt before and after the removal of the foreground objects to create a weakly-supervised \zerowastew~dataset. This dataset contains $1202$ frames with the foreground objects (\textit{before} class) and $1208$ frames without the foreground objects (\textit{after} class). One advantage of such a setup is that it is relatively cheap to acquire the ground truth labels (only an image-level inspection is required to ensure there are no false negatives in the \textit{after} class subset). The \zerowastew~dataset is specifically collected to be used in the weakly-supervised setup and is meant to provide an alternative and more data-efficient solution to the problem. The ground truth instance segmentation is available for all images of the \textit{before} class as it overlaps with the \zerowastef~dataset. Please see Figure~\ref{fig:data_collection_process} for examples of \zerowastew~dataset.

%% file: 4_experiments.tex
\section{Experiments}
\label{sec:experiments}
\begin{table}[t]
    \centering
    \inctabcolsep{-2.pt} {
    \begin{tabular}{r|cccccc} 
    \toprule
          & AP & AP50 & AP75 & APs & APm & APl  \\ 
         \midrule
         \textbf{\textit{RetinaNet}} & $21.0$ & $33.5$ & $ 22.2$ & $4.3$ & $9.5$ & $22.7$\\
         \textbf{\textit{MaskRCNN}} & $22.8$ & $34.9$ & $24.4$ & $4.6$ & $10.6$ & $25.8$ \\
         \textbf{\textit{TridentNet}}  & $\mathbf{24.2}$ & $\mathbf{36.3}$ & $\mathbf{26.6}$ & $\mathbf{4.8}$ & $\mathbf{10.7}$ & $\mathbf{26.1}$ \\
         \bottomrule
    \end{tabular}
    }
    \vspace*{-0.5em}
    \caption{Mean average precision on the test set of \zerowastef~ of MS-COCO-pretrained Mask R-CNN, TridentNet and RetinaNet finetuned on \zerowastef~. Please refer to Table~\ref{tab:coco_maskrcnn_classes} and Figure~\ref{fig:maskrcnn_res_examples} in the Appendix for class-wise results.
    \vspace{-5pt}
    }
    \label{tab:maskrcnn_res}
\end{table}
In this section, we provide baseline results for our proposed \zerowaste~dataset. We perform fully-supervised object detection on \zerowastef~using the most widely used Mask R-CNN~\cite{massa2018mrcnn}, RetinaNet~\cite{lin2017focal}, TridentNet~\cite{li2019scale} models, and we train DeepLabV3+~\cite{deeplabv3plus2018} as a semantic segmentation baseline. We also perform fully- and semi-supervised semantic segmentation on \zerowastes~using the ReCo~\cite{reco_ref} and CCT~\cite{Ouali_2020_CVPR}, and report the weakly-supervised segmentation results of RISE~\cite{petsiuk2018rise} CAMs, PuzzleCAM~\cite{puzzle_ref} and EPS~\cite{lee2021railroad} trained on \zerowastew.
We will provide the implementation and a detailed description of our experiments in the supplentary material.
\begin{table}[t]\small
    \centering
    \inctabcolsep{-3.3pt} {
    \begin{tabular}{r | c | c c | c c }
    \toprule
          \multirow{2}{*}{\textbf{Method}}
          &\multirow{2}{*}{\textbf{Supervision}} & \multicolumn{2}{c|}{\textbf{Validation}}  & \multicolumn{2}{c}{\textbf{Test}} \\
          &  & mIoU & Pixel Acc. & mIoU & Pixel Acc. \\ \midrule
        \multicolumn{6}{c}{Trained with \zerowastew}\\
        \midrule
        \textit{\textbf{Random}} & none & $7.2$ & $75.3$ & $8.4$ & $71.8$\\
        \textit{\textbf{CAM}} & \emph{weak-w} &  $16.7$ & $31.4$ & $19.1$ & $33.7$\\
        \textit{\textbf{PuzzleCAM}} & \emph{weak-w} & $29.87$ & $67.63$ & $28.46$ & $65.96$ \\
        \textit{\textbf{EPS}} & \emph{weak-w} & $34.56$ & $57.98$ & $34.06$ & $57.20$ \\
        \midrule
        \multicolumn{6}{c}{Trained with \zerowastef~and \zerowastes}\\
        \midrule
        
        \textit{\textbf{DeepLabv3+}} & \emph{full} & $46.93$ & $90.62$ & $52.13$ & $91.38$ \\
        \textit{\textbf{DeepLabv3+}} & \emph{full+aug} &  $47.48$ & $90.69$ & $\textbf{52.50}$ & $91.44$ \\
        \textit{\textbf{ReCo}} & \emph{full} &  $51.30$ & $89.22$ &  $52.28$ & $89.33$  \\
       \textit{\textbf{ReCo}}  & \emph{semi}  & $49.49$ & $89.58$ & $44.12$ & $88.36$ \\
        \textit{\textbf{CCT}} & \emph{full} & $30.79$ & $84.80$ & 
        $29.32$ & $85.91$ \\
        \textit{\textbf{CCT}} & \emph{semi} &  $28.70$ & $86.33$ & $32.49$ & $87.30$\\
       
       \textit{\textbf{EPS}} & \emph{weak-f} & $13.75$ & $59.96$ & $13.91$ & $60.65$ \\
        \bottomrule
    \end{tabular}
    }
    \caption{Results on validation and test sets of the \zerowastef~dataset of fully-, semi-, and weakly- supervised learning methods, separately trained on original non-augmented datasets and the augmented \zerowasteaug~dataset. `\emph{weak-w}', `\emph{weak-f}', and `\emph{weak-aug}' denote weak image-level supervision from \zerowastew~(before/after labels), \zerowastef~(class-wise labels), and \zerowasteaug~(class-wise labels) respectively. More details can be found in section~\ref{sec:sup_segm_res} of Appendix.}
    \vspace{-3pt}
    \label{tab:deeplab_res}
\end{table}
\subsection{Object Detection}
\label{sec:exp_segm}

\noindent \textbf{Experiments}
It has been shown that pretraining the model on a large-scale dataset, such as MS COCO~\cite{lin2014microsoft}, improves generalization and helps to prevent overfitting~\cite{huh2016makes, shin2016deep, girshick2014rich}. Therefore, in our first experiments, we used the model with weights learned on COCO and further fine-tuned it with our \zerowastef~dataset. 
We used a standard implementation of the popular Mask R-CNN, RetinaNet~\cite{lin2017focal} and the TridentNet~\cite{li2019scale} models both with ResNet-50~\cite{he2016deep} backbone. We used the Detectron2~\cite{wu2019detectron2} library in all of the experiments The model was finetuned for $40000$ iterations on the training set of our \zerowastef~dataset on a single Geforce GTX 1080 GPU with batch size $8$. To compensate for a relatively small number examples in the training set and to avoid overfitting, we leveraged heavy data augmentation, including random rotation and cropping, adjustment of brightness and hue, \etc 
 We report the experimental results in Table~\ref{tab:maskrcnn_res} (COCO $\rightarrow$ \zerowaste~section). A more detailed description of the results can be found in Section~\ref{sec:maskrcnn_exp} of Appendix.  

\paragraph{Results and Analysis}
{\color{black} The experimental results of Mask R-CNN, TridentNet ans RetinaNet indicate that the proposed \zerowaste~dataset is significantly challenging for the state-of-the-art detection methods, with TridentNet performing slightly better than other methods. All methods especially struggle with small objects, which are usually harder to label correctly. Additional results of the models pretrained on TACO dataset with labels mapped to \zerowaste~set of labels show that finetuning from a TACO-pretrained model slightly improves the detection precision (see Table~\ref{tab:coco_maskrcnn_classes} in the Appendix). } 
Recalling the history of success with other complex segmentation and detection datasets (\eg from mIoU $57\%$ in 2015~\cite{badrinarayanan2017segnet} to $84\%$ in 2020~\cite{cheng2020panoptic} on CityScapes~\cite{cordts2016cityscapes},
or from AP50 $29.9$~\cite{lin2014microsoft} in 2015 to $61.3$~\cite{xu2021end} on MS-COCO~\cite{lin2014microsoft} in 2021,
and knowing that the task \emph{can} be solved by humans with a little training, we believe that the computer vision community will eventually come up with efficient methods for this challenging task.
\subsection{Semantic Segmentation}

\begin{figure*}[ht]
    \centering
    \includegraphics[width=0.33\textwidth]{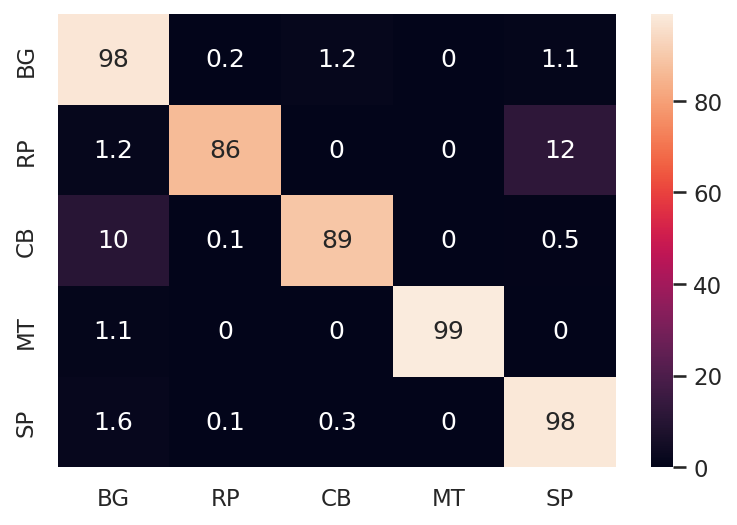}
    \includegraphics[width=0.33\textwidth]{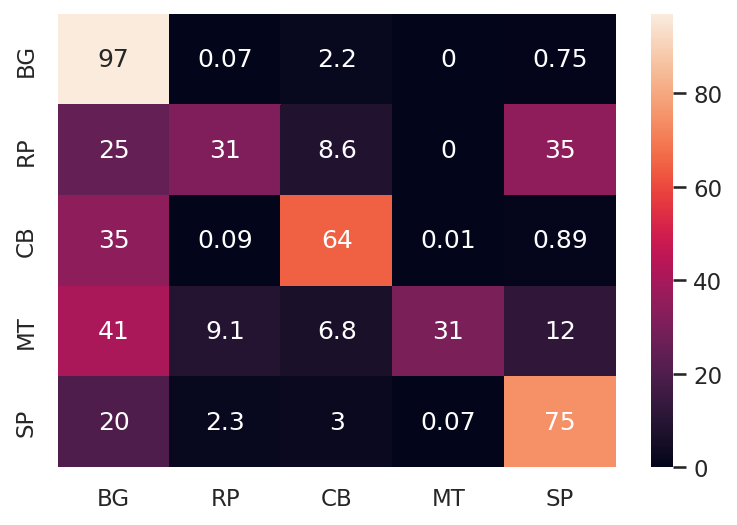}
    \includegraphics[width=0.33\textwidth]{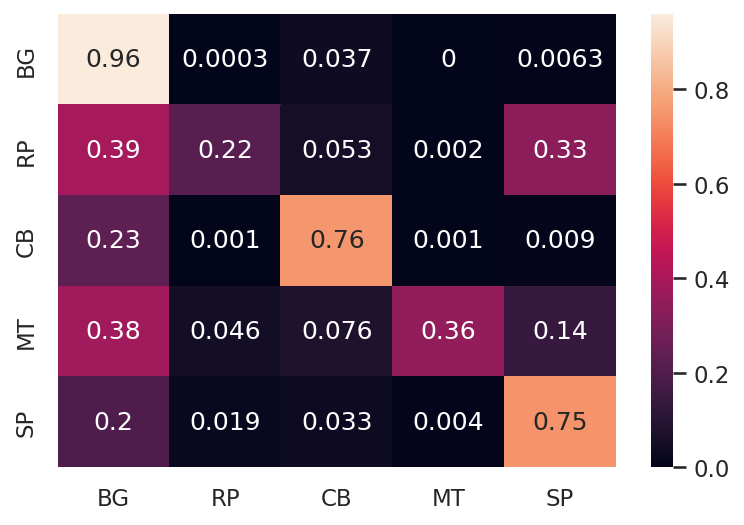}
    \caption{\textbf{Left:} Semantic segmentation confusion matrix of the manual annotations over $20$ frames represents human expert agreement after expert review, with the average pixel-level agreement above $94\%$ (see Figure~\ref{fig:conf_mtr_anno_noreview} of the Supplementary for the corresponding confusion matrix \emph{before} expert review). {\textbf{Center:} Confusion matrix of DeepLabV3+ on the test split of \zerowastef~. } {\textbf{Right:} Confusion matrix of DeepLabV3+ on the test split of \zerowastef~ when trained on~\zerowasteaug. }}
    \label{fig:conf_mrt_main}
\end{figure*}
\noindent\textbf{Fully-supervised Experiments} We used the state-of-the-art DeepLabv3+~\cite{deeplabv3plus2018} model as a fully-supervised semantic segmentation baseline for our dataset. DeepLabv3+ combines the atrous convolutions to extract features in multiple scales with an encoder-decoder paradigm to gradually sharpen the object boundary using the intermediate features. We used the model with ResNet-101 backbone with three $3\times3$ convolutions, froze the first three stages of the backbone, and separately fine-tuned the model on the training sets of \zerowastef~and \zerowasteaug. 

\noindent\textbf{Semi-supervised Experiments} 
For a semi-supervised segmentation baseline, we used the official
implementation of Regional Contrast~\cite{reco_ref} (ReCo) and CCT~\cite{Ouali_2020_CVPR} methods. ReCo utilizes the pixel-level local context and the global context represented in the semantic class relationships across the entire dataset, and the conventional Mean Teacher Framework in which a teacher network is used to generate pseudo-labels for the unlabeled images. CCT uses cross-consistency training that forces the predicted segmentation masks of unlabeled examples produced by decoders with various types of augmentations to be consistent. For comparison, we conducted four experiments with ReCo and CCT on \zerowastes~ in both the fully-supervised and semi-supervised learning settings. We used the default hyperparameters for both methods.

\paragraph{Weakly-supervised Experiments} 
For weakly-supervised segmentation, we followed the standard procedure  based on generating Class Activation Maps (CAMs) to serve as pseudo-labels for training a segmentation network. To do so, we conducted three experiments in which we trained a classifier using image-level labels from 1)  \zerowastew~(before/after labels) and 2) \zerowastef~(multiclass labels). For CAM generation, we incorporated the current state-of-the-art Puzzle-Cam~\cite{puzzle_ref} and EPS~\cite{lee2021railroad} methods that enhances CAM object coverage by matching the CAM of an entire image with those produced for the non-overlapping tiles of the image. We then threshold the generated CAMs with $0.45$ to produce pseudo segmentation labels for training the DeepLabv3+ segmentation network. 

\noindent\textbf{Results and Analysis}
Experimental results for fully-, semi- and weakly- supervised experiments are presented in Table~\ref{tab:deeplab_res}. Class-wise results and visualizations can be found in Section~\ref{sec:sup_segm_res} of the Appendix. {\color{black} Table~\ref{tab:deeplab_res} results indicate that our in-the-wild \zerowaste~dataset proposes a challenging semantic segmentation task. The confusion matrices on Figure~\ref{fig:conf_mrt_main} (center) show that the fully-supervised DeepLabV3+ can distinguish cardboard from paper in only $58\%$ of the cases, and missclassifies most rigid plastic objects as soft plastic. Additionally, metal objects are often segmented as background as they appear rarely in the original dataset. Further, training on \zerowasteaug~ substantially improves the pixel accuracy of the rare metal class on the original non-augmented test split of \zerowastef~(see Figure~\ref{fig:conf_mrt_main}, right), but reduces the performance on the rigid plastic. This suggests that instance-level data augmentation can be an efficient technique for semantic segmentation that reduces the generalization gap on the class-imbalanced data. 
The semi-supervised learning results indicate that the unlabeled examples from the \zerowastes~subset significantly worsened the performance of ReCo. This suggests that, while the inductive bias of the ReCo model that efficiently utilizes the object context information (\ie~surrounding background pixels) is not efficient for~\zerowaste~that has little to no context prior. In contrast, CCT performance increases slightly when the unlabeled examples are added during training, which suggests that the contrastive approach is more efficient for the proposed dataset. Additionally, results of the weakly-supervised experiments show that a simple CAM-based approach, as well as the state-of-the-art weakly supervised PuzzleCAM with cheap image-level annotations provide meaningful object localization cues. }

%% file: 5_discussion.tex
\section{Impact and Limitations of \zerowaste}
\label{sec:impact}

\noindent\textbf{Societal Impact}
We believe that human-robot collaboration is essential
 for more efficient computer-aided recycling, quality control of the sorting process, as well as in establishing safer work conditions for the MRF workers, \eg by detecting dangerous waste items.
In the big picture, establishing a human-robot workflow that can maximize efficiency, profit, safety and work quality in MRFs is an essential need to address the critical waste problem in an equitable and just manner.

\noindent\textbf{Limitations and Future Directions}
Despite the fact that \zerowaste~is the largest public dataset for waste detection to date, it is still smaller than the standard large-scale benchmarks due to the fact that the annotation process for this domain is very expensive. 
As future work, we plan to increase the dataset diversity by 
using synthetic-to-real domain adaptation and other data augmentation techniques.
\noindent\textbf{Acknowledgements} This paper was supported in part by the National Science Foundation under grant FW-HTF-RL \#1928506.

\section{Conclusion}
This work introduces \zerowaste~, which is the largest public dataset for waste detection up to date. \zerowaste~is designed as a benchmark for the training and evaluation of fully, weakly, and semi-supervised detection and segmentation methods, and can be used for various other tasks including label-efficient learning. We provide baseline results for the most popular fully, weakly, semi-supervised, and transfer learning techniques. Our results show that current state-of-the-art detection and segmentation methods cannot efficiently handle this complex in-the-wild domain. We anticipate that our dataset will motivate the computer vision community to develop more data-efficient methods applicable to a wider range of real-world computer vision problems.

%% file: 6_supplementary.tex
\clearpage
\section{Appendix}
\label{sec:appendix}

\subsection{Detection Experiments}
\label{sec:maskrcnn_exp}
\begin{figure*}
    \centering
    \includegraphics[width=1.\linewidth]{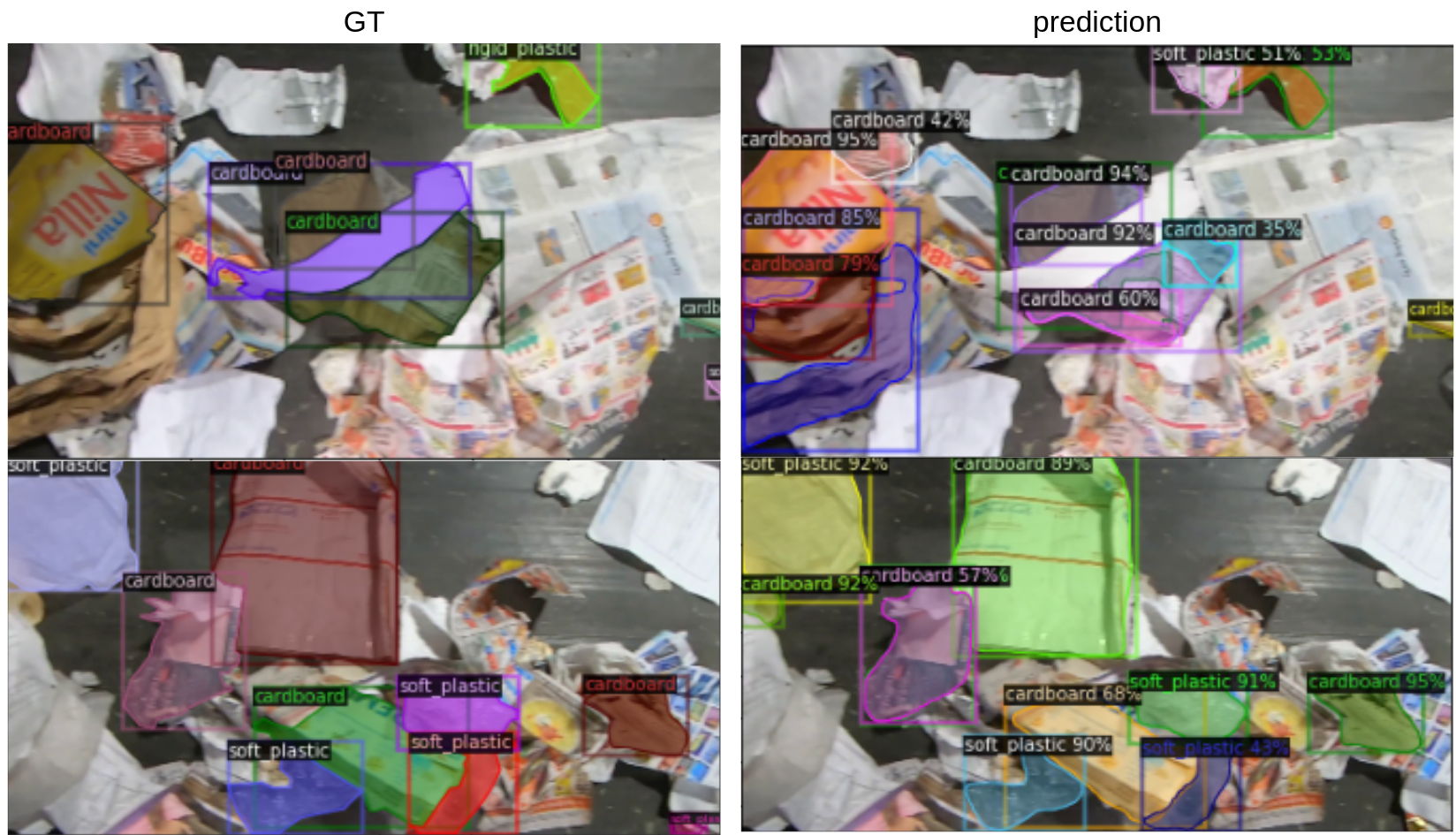}
    \caption{Examples of predictions of Mask-RCNN trained on \zerowastef. }
    \label{fig:maskrcnn_res_examples}
\end{figure*}

In this section, we provide detailed class-wise results of object detection experiments using Mask R-CNN. Please see Table~\ref{tab:coco_maskrcnn_classes} for detailed results of the experiments with Mask R-CNN pretrained with COCO and TACO datasets, and the visualization of the Mask-RCNN predictions can be found on Figure~\ref{fig:maskrcnn_res_examples}.

\paragraph{Implementation details} 
For Mask R-CNN and TridentNet experiments, we used the default implementation and hyperparameters from the Detectron V2~\cite{wu2019detectron2} library. For the RetinaNet experiments, we used the open-source implementation and the default hyperparameters provided here:\url{https://github.com/yhenon/pytorch-retinanet}. We provide the exact code in the supplementary files.

\begin{table}[ht]
    \centering
    \inctabcolsep{-2.pt} {
    \begin{tabular}{c|c c | c c }
    \toprule
    & \multicolumn{2}{c|}{\textbf{COCO}} & 
    \multicolumn{2}{c}{\textbf{TACO}} \\
    
    & \textit{Validation} & \textit{Test} &  \textit{Validation} & \textit{Test} \\ 
    \midrule
     \textit{Cardboard} & $29.85$ & $35.05$ & $ 30.50$ & $36.22$ \\
     \textit{Soft Plastic} & $22.09$ & $26.29$ & $ 22.96$ & $28.28$ \\
     \textit{Rigid Plastic} & $15.70$ & $14.57$ & $16.58$ & $16.24$ \\
     \textit{Metal} & $1.64$ & $11.97$ & $5.20$ & $16.53$ \\
     \textit{Total} & $17.32$ & $21.97$ & $18.81$ & $24.32$ \\
    \bottomrule
    \end{tabular}
    }
    \vspace*{0.5em}
    \caption{Class-wise average precision results of a COCO-pretrained (\textbf{left}) and TACO-pretrained (\textbf{right}) Mask R-CNN on our \zerowastef~dataset. }
    \label{tab:coco_maskrcnn_classes}
\end{table}

\subsection{Segmentation Experiments}
\label{sec:sup_segm_res}

\begin{figure*}
    \centering
    \includegraphics[width=1.\linewidth]{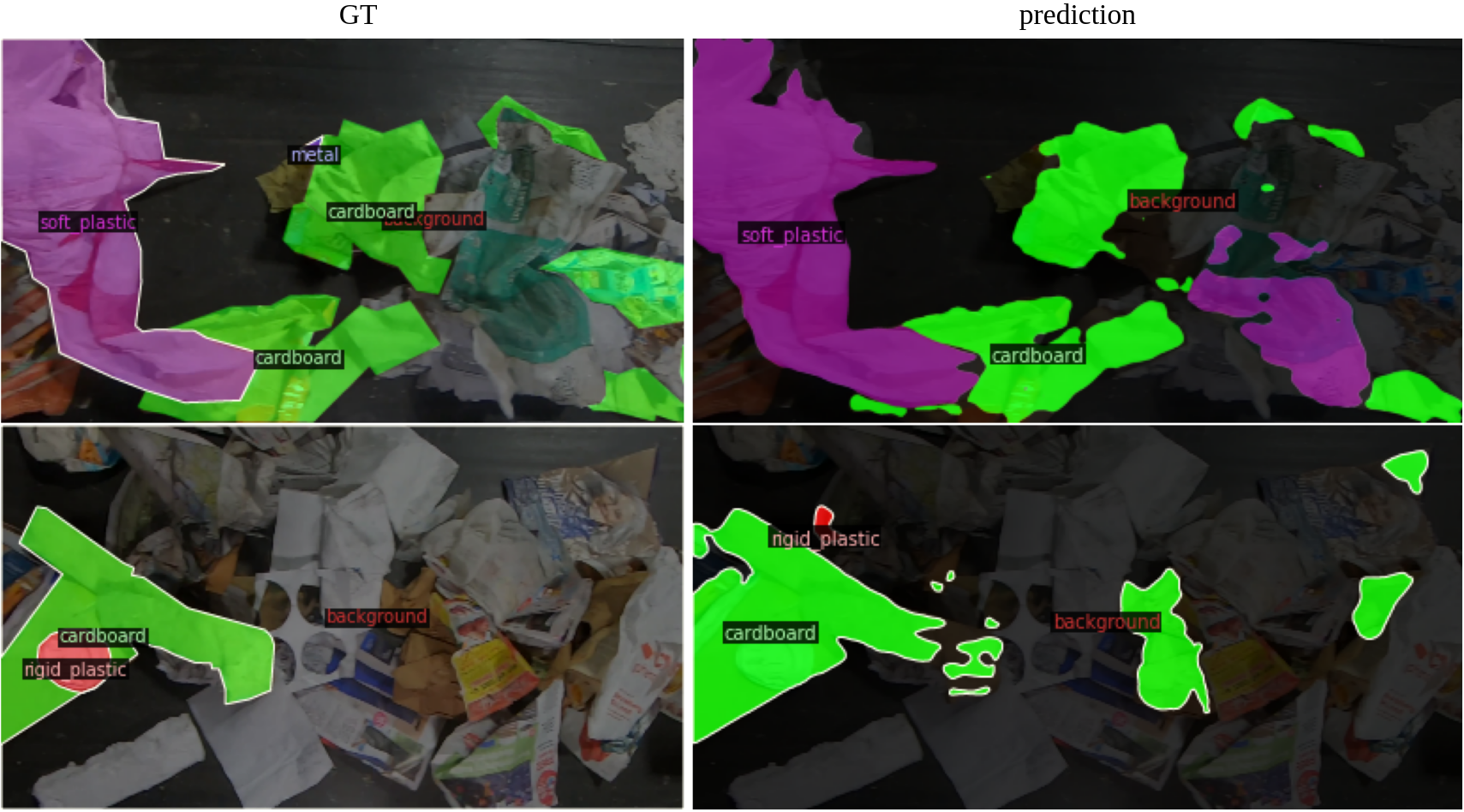}
    \includegraphics[width=1.\linewidth]{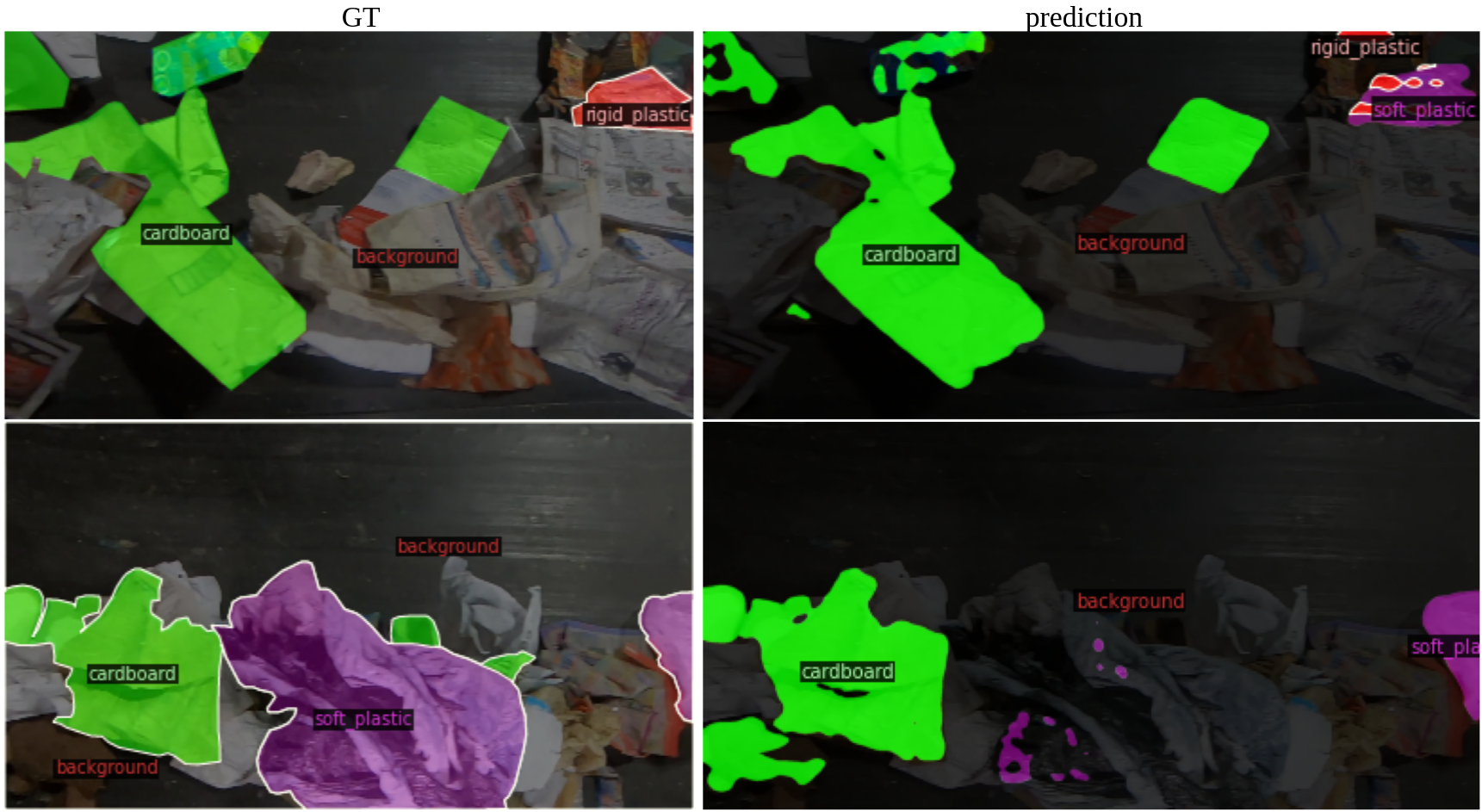}
    \caption{Examples of predictions of the fully-supervised DeepLabv3+  method trained on \zerowastef. }
    \label{fig:deeplab_res_examples}
\end{figure*}

\begin{figure*}
    \centering
    \includegraphics[width=.8\textwidth]{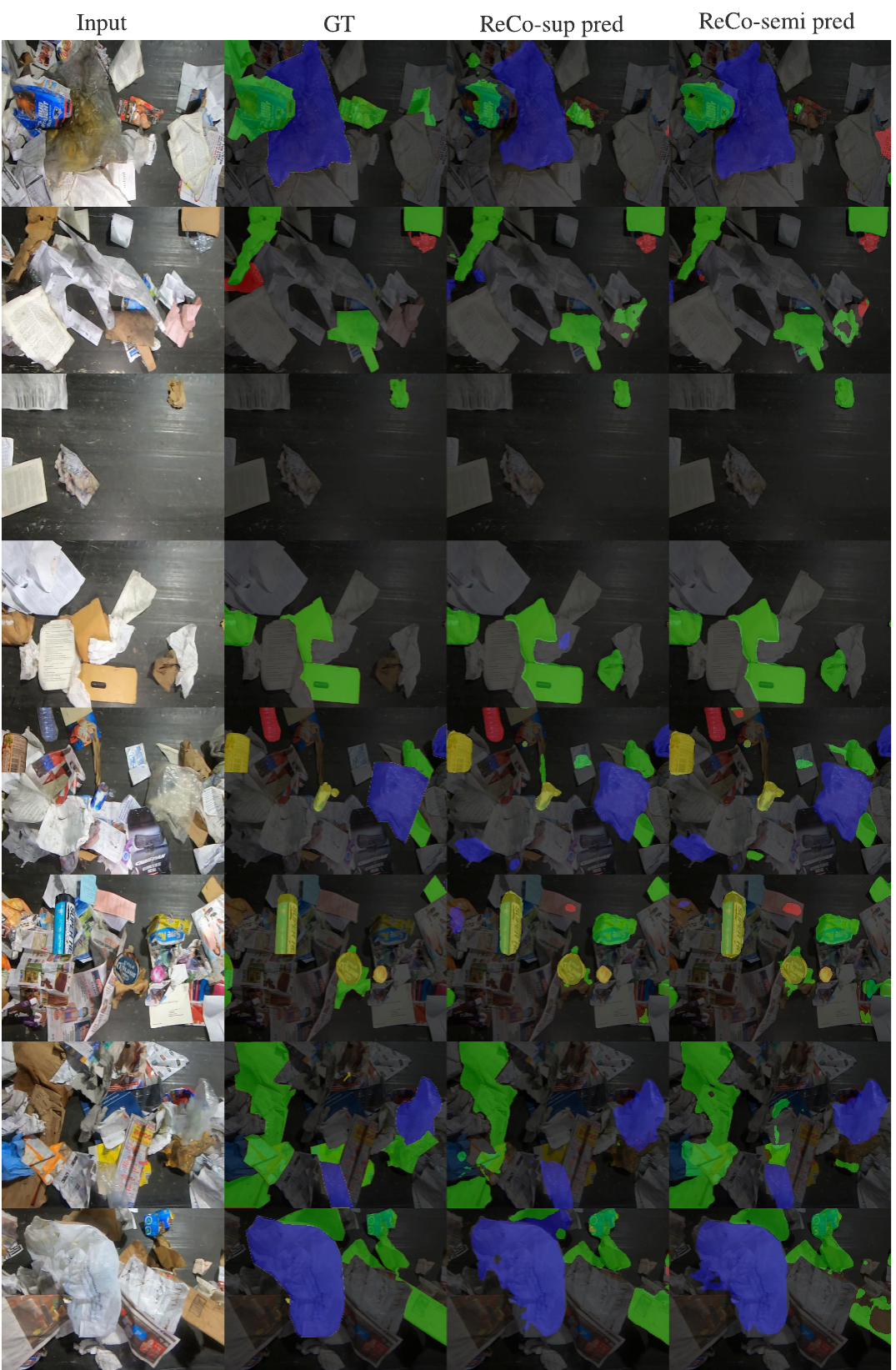}
    \caption{Examples of predictions of the supervised and semi-supervised versions of the ReCo method on the images from the validation set. }
    \label{fig:reco_res_examples}
\end{figure*}

\begin{figure*}
    \centering
    \includegraphics[width=0.7\textwidth]{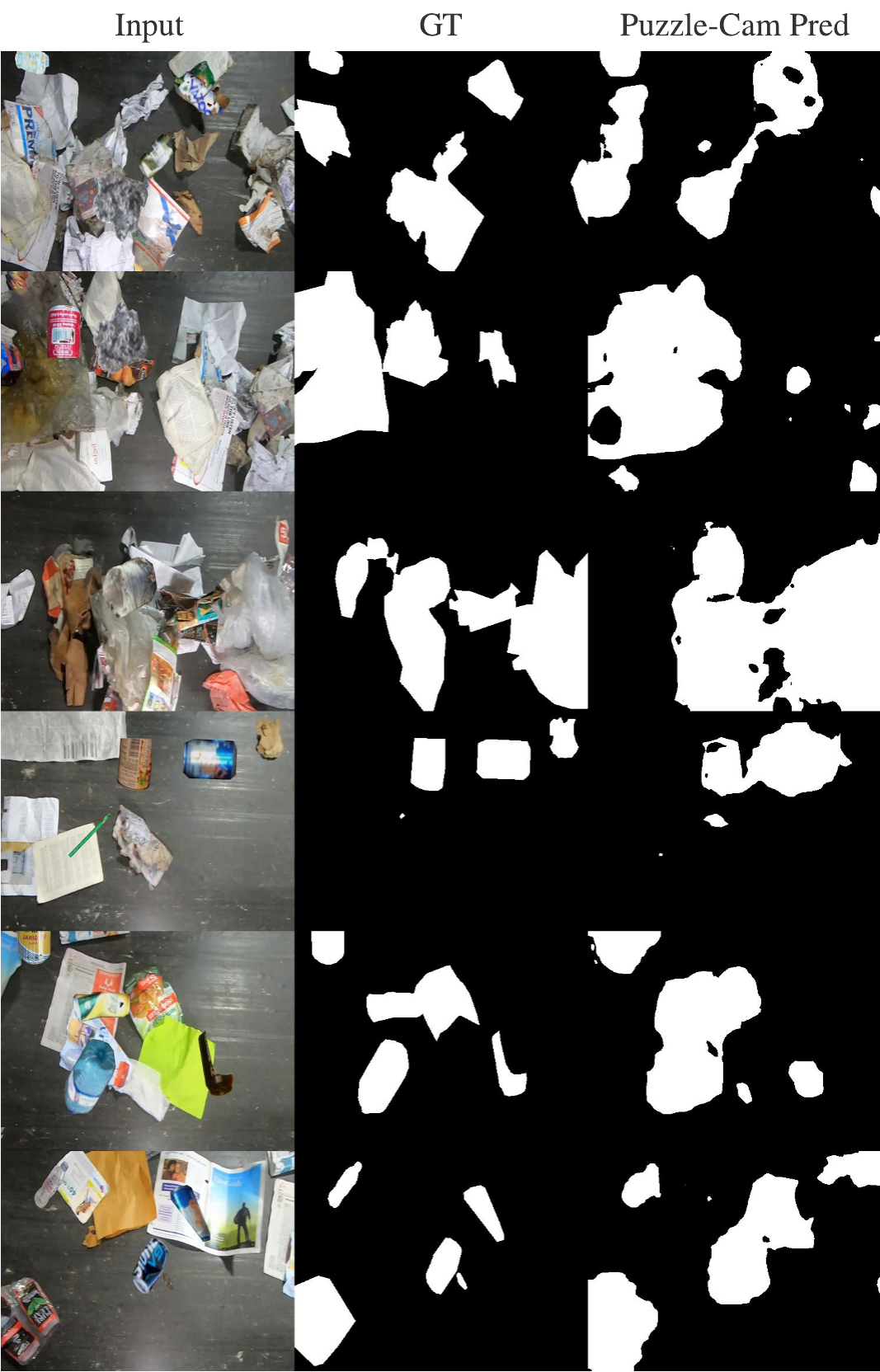}
    \caption{Examples of predictions of the weakly-supervised Puzzle-Cam trained on \zerowastew~(before/after image-level labels).}
    \label{fig:puzzle_res_examples}
\end{figure*}

In this section, we provide detailed class-wise results, confusion matrices, and visualizations of our segmentation experiments. Table~\ref{tab:sup_deeplab_res} shows the segmentation results of DeepLabv3+~\cite{deeplabv3plus2018} (trained on \zerowastef) for each class in \zerowaste~dataset.  Visual predictions of DeepLabv3+, ReCo, and Puzzle-Cam~\cite{puzzle_ref}, all trained on \zerowasteaug, can be found in Figures~\ref{fig:deeplab_res_examples},~\ref{fig:reco_res_examples}, and~\ref{fig:puzzle_res_examples} respectively.

\paragraph{Implementation details}
For all semantic segmentation experiments, we used the default hyperparameters for all methods. We used the Detectron V2~\cite{wu2019detectron2} implementation for the DeepLabv3+ experiments, as well as the official implementation of ReCO (\url{https://github.com/lorenmt/reco}), CCT (\url{https://github.com/yassouali/CCT}), PuzzleCAM (\url{https://github.com/OFRIN/PuzzleCAM/}) and EPS (\url{https://github.com/neo85824/epsnet}). 

\begin{table*}[hb]
    \centering
    \begin{tabular}{r | c c c | c c c | c c c}
    \toprule
          & \multicolumn{3}{c}{\textbf{Train}}  & \multicolumn{3}{|c|}{\textbf{Validation}}  & \multicolumn{3}{c}{\textbf{Test}} \\
          
          & IoU & Precision & Pix. Acc. & IoU & Precision & Pix. Acc. & IoU & Precision & Pix. Acc. \\ 
          \midrule
          
        \textit{Background} & 
        $94.82$ & $96.31$ & $98.40$
        & $90.41$ & $92.81$ & $97.22$ 
        & $91.02$ & $ 93.71$ & $ 96.95$ \\
        \textit{Cardboard} &
        $69.51$ & $89.21$ & $75.86$ 
        & $51.38$ & $79.96$ & $58.98$
        & $54.47$ & $ 77.95$ & $ 64.40$ \\
        \textit{Soft plastic} & 
        $78.99$ & $91.09$ & $85.61$
        & $61.86$ & $77.34$ & $75.56$
        & $63.18$ & $80.99$ & $74.17$ \\
       \textit{Rigid plastic} & 
       $65.46$ & $76.21$ & $82.27$ 
       & $27.58$ & $59.33$ & $34.01$ 
       & $24.82$ & $54.95$ & $31.16$ \\
        \textit{Metal} &  
        $76.23$ & $86.78$ & $86.98$ 
        & $3.41$ & $33.19$ & $3.67$ 
        & $27.14$ & $70.54$ & $30.61$ \\
        \midrule
        
        \textbf{mean} &
        $77.01$ & $87.96$ & $85.64$
        & $46.93$ & $83.35$ & $53.88$
        & $52.13$ & $75.63$ & $59.46$ \\
        \bottomrule
    \end{tabular}
    \vspace*{0.5em}
    \caption{\footnotesize{Experimental results of DeepLabv3+~\cite{deeplabv3plus2018} on our \zerowastef~dataset.}}
    \label{tab:sup_deeplab_res}
\end{table*}

\subsection{Additional Data Examples}
\label{sec:more_examples}
\begin{figure*}[hb]
    \centering
    \textbf{Original Frame} \hspace{4cm} \textbf{Processed Frame}
    \includegraphics[width=\textwidth]{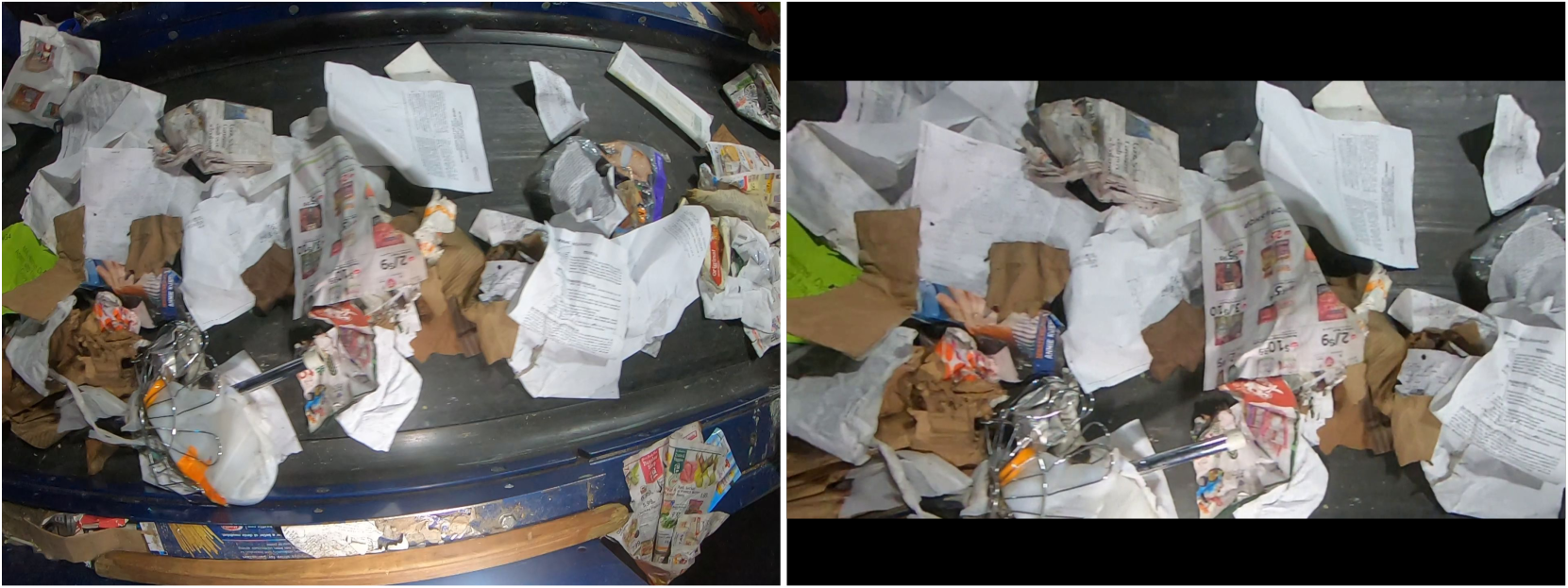}
    \caption{\textbf{Left:} sample video frame from \zerowaste~\textit{before} collection of target objects to be removed from the conveyer belt. \textbf{Right:} the same video frame processed as described in section~\ref{sec:dataset_description}: fisheye effect removed, frame rotated to make the conveyor belt parallel to the image border, regions outside conveyor belt cropped out, motion blur removed.}
    \label{fig:processing}
\end{figure*}

\begin{figure*}[hb]
    \centering
    \includegraphics[width=1.\textwidth]{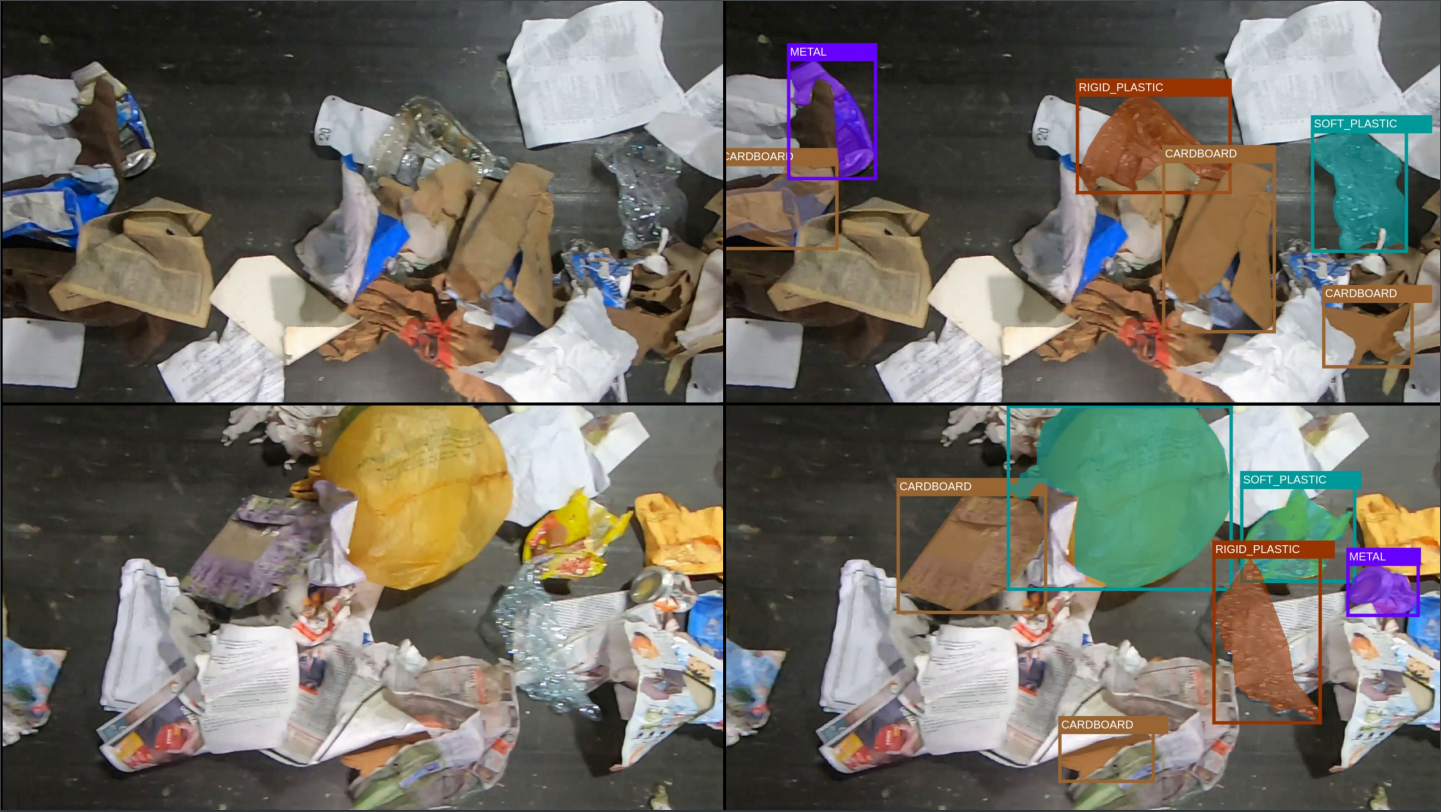}
    \caption{Examples of images (\textbf{left}) and the corresponding polygon annotation (\textbf{right}) of the proposed \zerowaste~dataset. }
    \label{fig:more_fully_annotated_examples_1}
\end{figure*}

\begin{figure*}[hb]
    \centering
    \includegraphics[width=1.\textwidth]{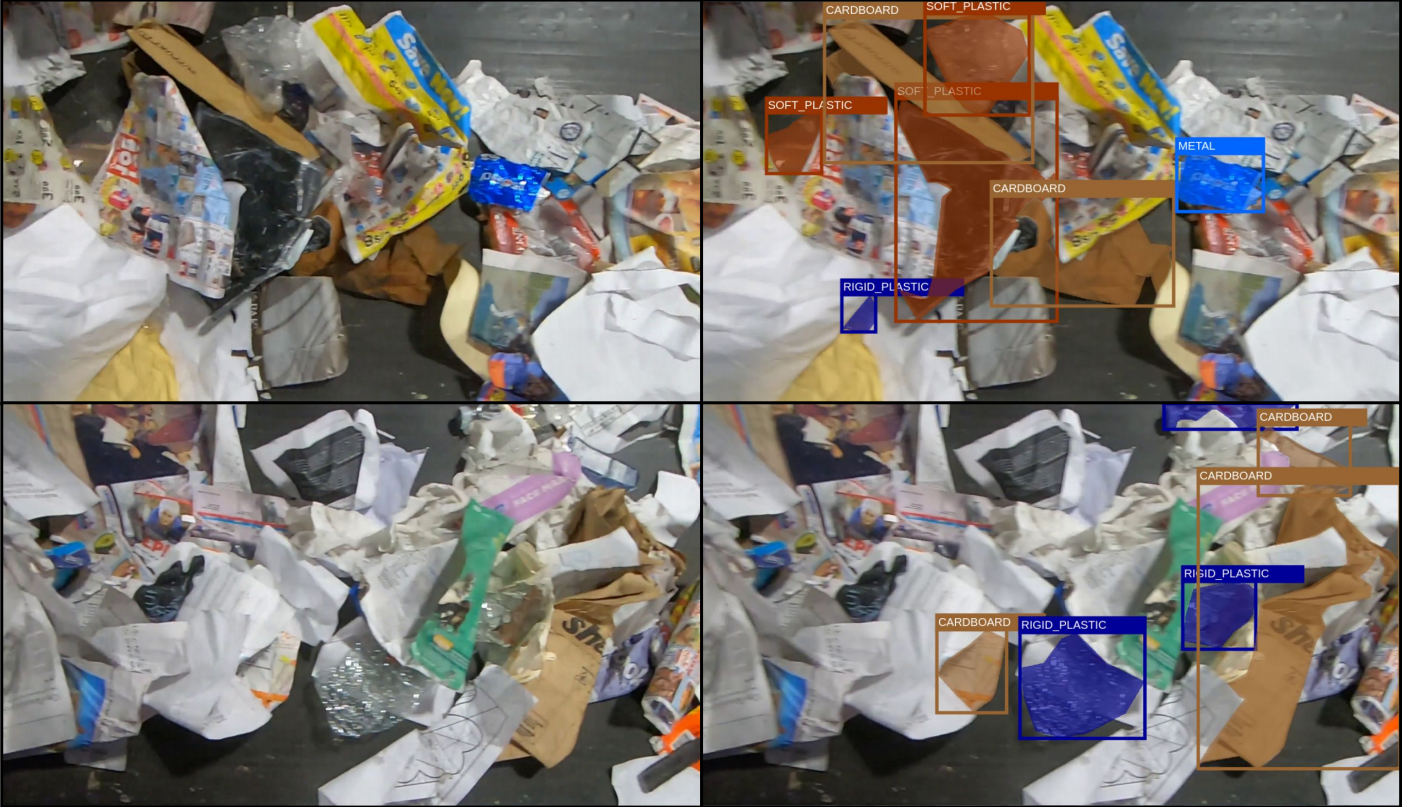}
    \caption{Examples of images (\textbf{left}) and the corresponding polygon annotation (\textbf{right}) of the proposed \zerowaste~dataset. }
    \label{fig:more_fully_annotated_examples_2}
\end{figure*}

\begin{figure*}[hb]
    \centering
    \includegraphics[width=1.\textwidth]{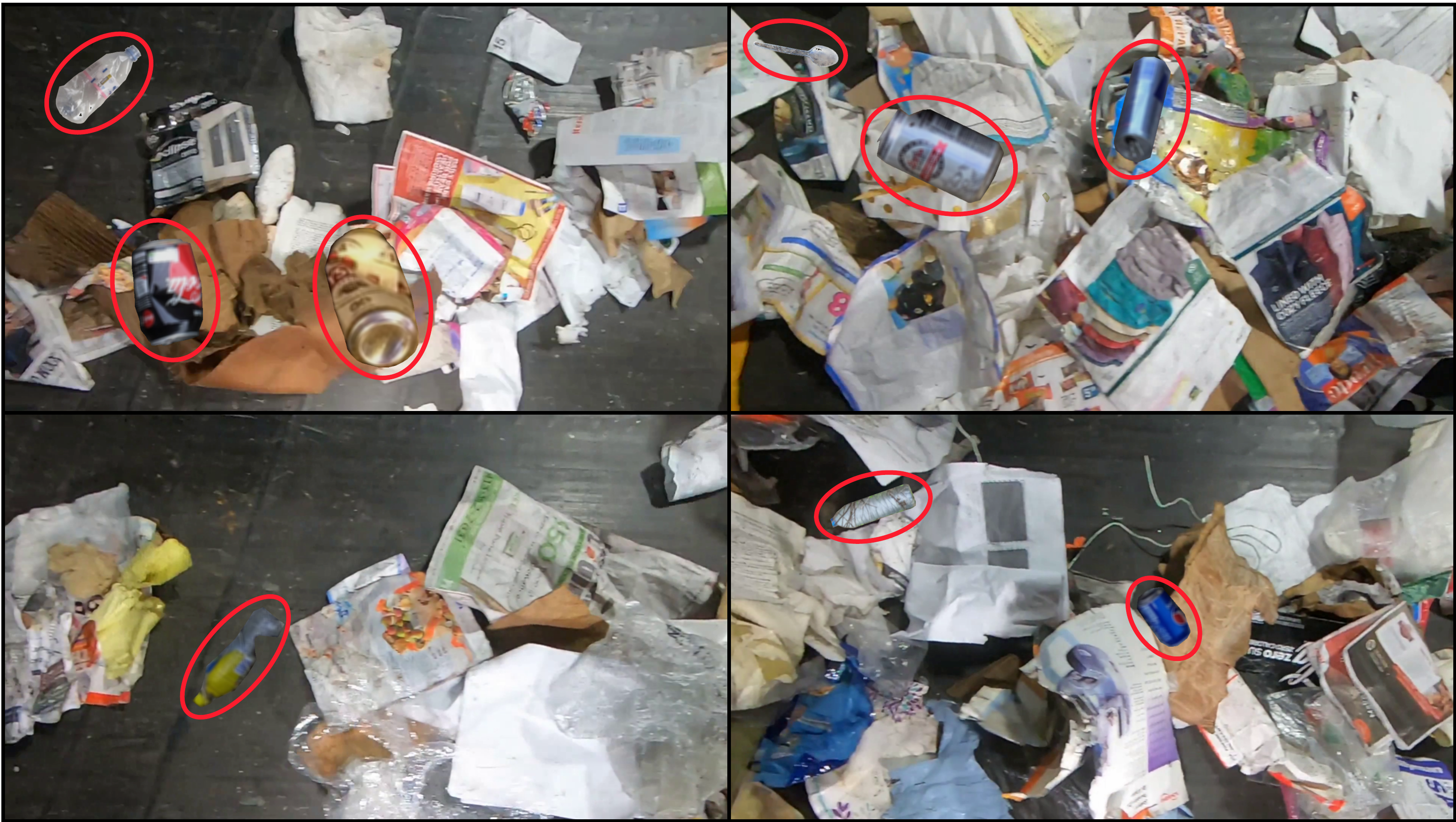}
    \caption{Examples of images from \zerowasteaug~. {\color{red} Red circles} show examples of metal and rigid plastic augmented objects cropped out from TACO.}
    \label{fig:more_fully_annotated_examples_1_aug}
\end{figure*}

In this section, we provide further examples of frames from our \zerowastef~and \zerowasteaug~datasets. The example of a frame from \zerowaste~before and after processing as described in Section~\ref{sec:dataset_description} can be found in Figure~\ref{fig:processing}. Examples of fully annotated \zerowastef~frames can be found in Figures~\ref{fig:more_fully_annotated_examples_1} and~\ref{fig:more_fully_annotated_examples_2}. Further, examples of augmented frames of \zerowasteaug~can be found in Figure~\ref{fig:more_fully_annotated_examples_1_aug}.

\subsection{Annotation Costs and Statistics}
\label{sec:anno_costs}
The total annotation cost per frame is $1.21\$$, including $0.6\$$ for annotation and  $0.62\$$ for expert review. 
\begin{figure}
    \centering
    \includegraphics{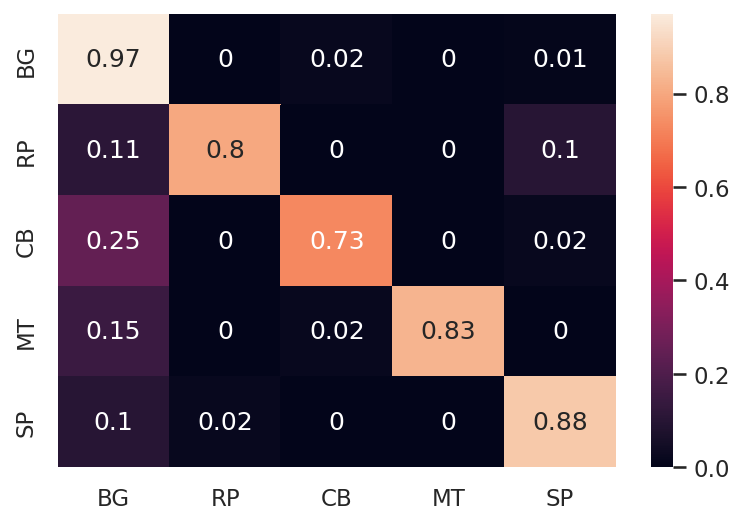}
    \caption{Confusion matrix of corresponding to average agreement across annotators over 20 frames \emph{before} the expert review.}
    \label{fig:conf_mtr_anno_noreview}
\end{figure}